\documentclass{article} 
\usepackage{iclr2026_conference,times}
\pdfoutput=1

\usepackage{amsmath,amsfonts,bm}

\def\eqref#1{equation~\ref{#1}}

\def\1{\bm{1}}

\DeclareMathAlphabet{\mathsfit}{\encodingdefault}{\sfdefault}{m}{sl}
\SetMathAlphabet{\mathsfit}{bold}{\encodingdefault}{\sfdefault}{bx}{n}

\usepackage{hyperref}
\usepackage{url}

\usepackage{graphicx}
\usepackage{booktabs, multirow}
\usepackage{textcomp}
\usepackage{gensymb}
\usepackage[acronym]{glossaries}
\glsdisablehyper

\usepackage{subcaption}
\usepackage{bm}

\usepackage{xspace}
\usepackage{cleveref}
\Crefname{section}{Sec.}{Secs.}
\Crefname{tabular}{Tab.}{Tabs.}
\Crefname{table}{Tab.}{Tabs.}
\Crefname{equation}{Eq.}{Eqs.}
\Crefname{figure}{Eq.}{Eqs.}

\usepackage{wrapfig}
\usepackage{float}
\usepackage{enumitem}
\usepackage{amsmath}
\usepackage{algorithm}
\usepackage{algpseudocode}

\newcommand{\algname}{SWM}

\newcommand{\algnamefull}{Semantic World Model}
\newcommand{\algnamefulls}{Semantic World Models}

\newcommand{\method}{SWM }
\newcommand{\methodfull}{Semantic World Model }
\newcommand{\methodfulls}{Semantic World Models }

\title{Semantic World Models}

\author{
Jacob Berg$^{1}$ \quad
Chuning Zhu$^{1}$ \quad
Yanda Bao$^{1}$ \quad
Ishan Durugkar$^{2}$ \quad
Abhishek Gupta$^{1}$ \\
\\
$^{1}$University of Washington \quad
$^{2}$Sony AI \\
\\
\texttt{\nolinkurl{{jacob33,zchuning,yandabao,abhgupta}@cs.washington.edu}} \\
\texttt{\nolinkurl{ishan.durugkar@sony.com}}
}

\newacronym{swm}{SWM}{Semantic World Model}
\newacronym{vqa}{VQA}{Visual Question Answering}
\newacronym{vlms}{VLMs}{Vision Language Models}

\iclrfinalcopy 
\begin{document}

\maketitle

\begin{abstract}
Planning with world models offers a powerful paradigm for robotic control. Conventional approaches train a model to predict future frames conditioned on current frames and actions, which can then be used for planning. However, the objective of predicting future pixels is often at odds with the actual planning objective; strong pixel reconstruction does not always correlate with good planning decisions. This paper posits that instead of reconstructing future frames as pixels, world models only need to predict task-relevant \emph{semantic} information about the future. For such prediction the paper poses world modeling as a visual question answering problem about semantic information in \emph{future frames}. This perspective allows world modeling to be approached with the same tools underlying vision language models. Thus, vision-language models can be trained as "semantic" world models through a supervised finetuning process on image-action-text data, enabling planning for decision-making while inheriting many of the generalization and robustness properties from the pretrained vision-language models. In this paper, we introduce this new method and validate it on simulated robotic tasks leading to improvements over typical paradigms of reconstruction-based action-conditional world modeling.
\vspace{0.5em}
\begin{center}
\url{https://weirdlabuw.github.io/swm}
\end{center}


\end{abstract}

\vspace{-1em}

\section{Introduction}
\vspace{-1em}

World models are a class of learning methods capable of absorbing large amounts of data to make generative predictions about future outcomes in the world. These predictions can then be used to inform decision-making via planning \citep{MPPI, Hafner2019PlaNet, Rybkin2021LatCo, hansen2022temporaldifferencelearningmodel}, helping policies acquire generalizable and robust behaviors. The practical instantiations of world models are diverse, ranging from smaller state-based dynamics models \citep{bo2025areview} to large action-conditioned video prediction models \citep{genie3}. Across these instantiations, pixel-level reconstruction of future observations is commonly used as a training recipe. While these approaches are often successful at generating realistic images, as evident from high-quality video generations, they can be challenging to use for planning. Despite the visual fidelity, these predictions often miss (or misrepresent) key semantic details necessary for decision making, e.g., the details of precise dexterous contact. 
While there have been suggestions for modeling "task-relevant" latent representations \citep{zhang2021learning, hansen2022temporaldifferencelearningmodel, zhu2023repo}, these methods often impose additional assumptions on the availability of rewards \citep{hansen2024tdmpc} or known factors \citep{locatello2020slot}, making them challenging to use in practice across a variety of world modeling problems. 

If pixels are not necessary for planning, what is actually needed to make decisions about acting in the world? This paper posits that the ability to predict \emph{semantic} information about future outcomes is sufficient. Rather than forecasting raw visual frames, world models should capture task-relevant information about objects and their interactions, e.g., “Did the arm get closer to the object?”, “Did the red cube tip over?”, “Was the blue moon picked up?”. This work frames such information as a visual question-answering (VQA) problem about the future, leveraging the fact that any desired outcome can be expressed as a set of yes/no questions\footnote{Other textual question-answer types may be applicable as well.}. That is, \textit{the problem of world modeling can be redefined as a VQA problem about outcomes in the future}.

There already exists a class of models with extensive tooling for VQA on static observations, i.e., vision-language models (VLMs). For world modeling, VLMs offer two key advantages: they provide a strong foundation for VQA through large-scale pretraining and broad generalization, and they encode prior knowledge about which tasks and semantic features are relevant in a scene. These strengths make frontier VLMs well suited to formulating task-relevant questions and producing reliable answers when given static observations. However, their lack of predictive capacity about future outcomes limits their direct utility for decision-making.

\begin{figure}[!t]
\vspace{-1em}
\centering
\includegraphics[width=\linewidth]{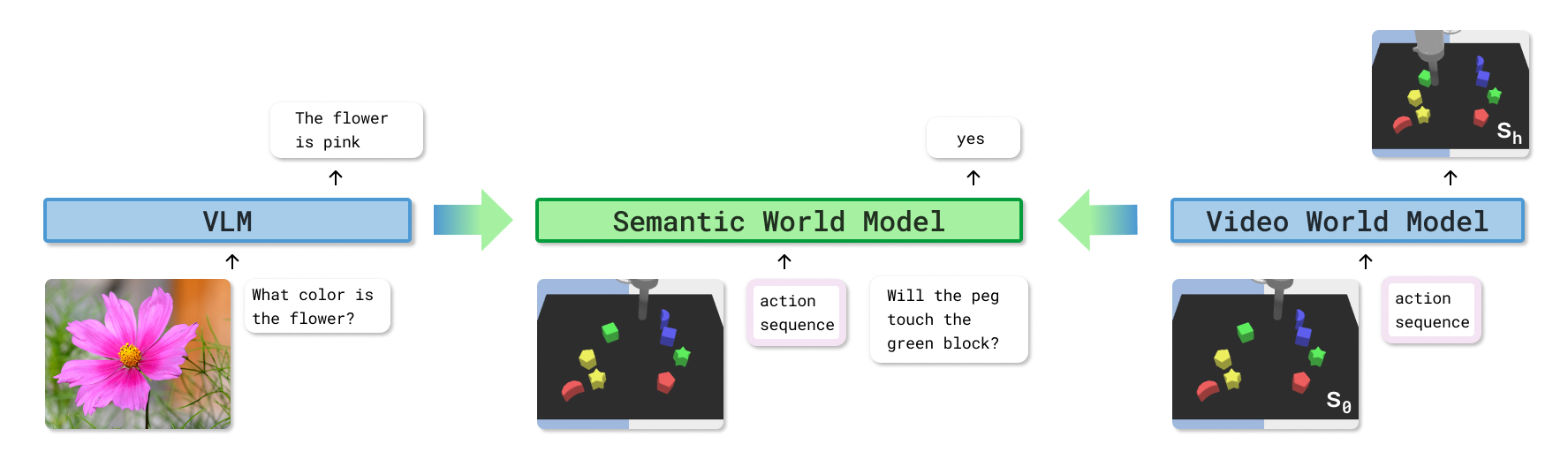}
\caption{\footnotesize{Comparison between Vision-Language Models, Video World Models, and \textbf{\algnamefulls{}}. While Vision-Language Models answer questions about static observations and Video World Models predict future observations given actions, \algnamefulls{} take observations and actions as input to directly answer questions about the future outcomes of those actions.}}
\label{fig:teaser}
\vspace{-1.8em}
\end{figure}

This work proposes a new paradigm for world modeling - Semantic World Models (SWM): a world model that is represented as an action-conditional vision-language model that answers questions about the semantic effects of actions in the future. Unlike traditional world models that predict future frames, a \algnamefull{} \textit{answers questions about the future} given current observations (represented as an image) and a sequence of actions. As shown in Fig. \ref{fig:teaser}, the model takes as input the current observations, a proposed action sequence, and a natural language query about the future. It then generates a textual answer by understanding the consequences of taking the actions in the environment. Since \algname{} is fundamentally a task-agnostic world model, it can be trained on general sequential data with minimal quality assumptions, including both play and suboptimal data. The training data can be easily obtained from any (expert or non-expert) data corpus in the format of current observations, actions, questions (about the future), and expected answers. 


The ability to reason about outcomes in the future with an SWM enables flexible open-world multi-task planning in action space: given a task specification in natural language, one could either leverage a pre-trained VLM \citep{GPT4o_2024, paligemma} or manually decompose the task specification into a set of questions and expected answers in text form. Given this QA set, \algname{} can then be used to plan actions that elicit the expected answers to these questions \textit{in the future} with high likelihood. While a plethora of techniques can be used for this planning, this work shows compatibility with both zero-order sampling-based methods \citep{cem, MPPI} and first-order gradient planning methods \citep{ruder2017overviewgradientdescentoptimization, Rybkin2021LatCo} that perform optimization with respect to the expected likelihood objective. It shows that these planning methods can be computationally tractable, enabling a significant test-time improvement over nominal action selection methods. Moreover, it demonstrates the extensibility of such planning methods to multi-step long-horizon problems. 

\algname{} is empirically evaluated on a suite of multiple different tasks in two commonly used multi-task simulation domains -- Language Table (LangTable) \citep{langtable} and OGBench \citep{ogbench_park2025}. This evaluation shows that (1) \algname{} can accurately answer questions about future outcomes while generalizing to novel scenes, and (2) \algname{} can be combined with standard sampling-based planning techniques and a gradient-based improvement technique to solve diverse robotics tasks with considerable policy improvement through test-time optimization.  \algname{} introduces a new class of world models that leverage the rich pretraining knowledge from VLMs for grounded, flexible, and scalable robotic control.
\section{Related Work}
\vspace{-.5em}

\paragraph{Vision-Language Models (VLMs)} broadly encompass representation learning methods and multimodal generative models trained on vision and language data. Representation learning methods jointly train a vision encoder and a text encoder by aligning their encoded representations. These representations can then be utilized in various applications, such as classification, retrieval, and control. CLIP \citep{radford21clip} learns such representations from image-text data by utilizing a contrastive loss, contrasting positive image-text pairs with negative pairs. SigLIP \citep{Zhai_2023_ICCV} replaces the contrastive loss with a pairwise sigmoid loss to facilitate scalable training. Multimodal generative models, commonly known as VLMs, enable a broad range of promptable behaviors such as understanding, summarizing, and question answering \citep{GPT4o_2024, gemini2023, Molmo_PixMo_2024, Qwen_2023, paligemma, LLaMA_2023}. A VLM takes in an image and a language prompt as input and generates a natural language response. They are typically trained with a next-token prediction objective. Recently, a family of vision-language-action models (VLAs) has been introduced to bring the vision-language understanding capabilities of VLMs to embodied decision-making \citep{BrohanBrownCarbajalEtAl_2023_RT1, KimPertschKaramchetiEtAl_2025_OpenVLA, BlackBrownDarpinianEtAl_2024_Pi0}. VLAs are trained on annotated robot trajectories to generate actions conditioned on image observations and language instructions. OpenVLA \citep{KimPertschKaramchetiEtAl_2025_OpenVLA} directly predicts discrete action tokens, while Pi-0 \citep{BlackBrownDarpinianEtAl_2024_Pi0} decodes actions via a diffusion action head. Unlike VLAs, an \algname{} takes in observations, actions, and a natural language prompt as input, and generates a natural language response about the future after taking the actions. In some sense, an \algname{} can be viewed as an ``inverted'' VLA, where the actions become the input and the language becomes the output. This approach hypothesizes that using language as the output format can better retain the pretraining knowledge of VLMs, since they were trained with next token prediction objectives.
\vspace{-.5em}

\paragraph{World Models for Control} are approximate models of the dynamics of the world, typically trained to predict future observations conditioned on current observations and actions. The ability to forecast the future without interacting with the world can greatly facilitate decision-making and control. A prominent line of work focuses on planning with world models. \citep{Chua2018PETS, Hafner2019PlaNet, Rybkin2021LatCo}. PETS \citep{Chua2018PETS} learns a one-step dynamics model and applies the cross-entropy method to plan for optimal actions for a given reward. PlaNet \citep{Hafner2019PlaNet} learns a recurrent latent dynamics model with a reconstruction objective and applies planning in the latent space. LatCo \citep{Rybkin2021LatCo} leverages collocation-based planning to enable long-horizon planning with latent dynamics models. Another line of work utilizes world models as a simulator for reinforcement learning \citep{Hafner2020Dream, zhang2021learning, hansen2022temporaldifferencelearningmodel}. Dreamer \citep{Hafner2020Dream} and TD-MPC \citep{hansen2022temporaldifferencelearningmodel} use a latent dynamics model to generate rollouts for actor-critic policy optimization, achieving remarkable sample efficiency. \citep{zhang2021learning} learns a latent representation predictive of dynamics and reward, which can then be used as an invariant representation for RL policies. Recently, world models have been used together with imitation learning methods to facilitate out-of-distribution generalization \citep{du2023learning, zhu2025uwm}. UniPi \citep{du2023learning} uses a world model as a high-level planner to condition low-level policies. UWM \citep{zhu2025uwm} trains a unified video-action diffusion model, incorporating video data into pretraining to improve generalization. Unlike these explicit world models, \algname{} understands the dynamics of the world by reasoning in language space, allowing the model to bootstrap from the Internet-scale pretraining of VLMs. \algname{} can then be used with planning techniques to derive versatile language-conditioned policies.

\section{World Modeling as VQA}

\begin{figure*}[ht]
    \vspace{-1em}
    \centering
    \includegraphics[width=.8\textwidth]{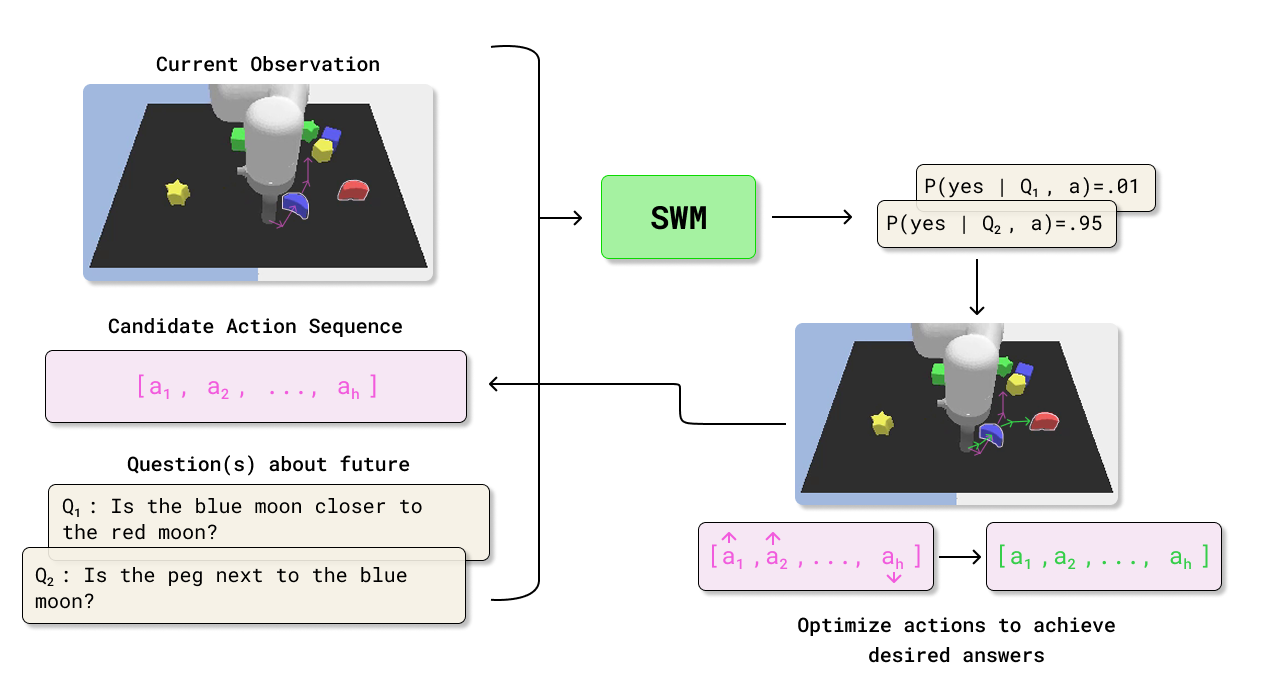}
    \caption{\footnotesize{\textbf{Overview of \algnamefulls.} SWM is a VLM adapted to answer questions about the future realized by the actions used to condition the model. Using a set of questions and desired answers, its predictions can be converted into a planning signal and iteratively refine the action sequence.}}
    \label{fig:methodfigure}
    \vspace{-1.5em}
\end{figure*}

This section presents details of the data generation pipeline, the SWM architecture, and the training methodology. It then touches on the sampling-based and gradient-based planning methods used for policy extraction under \algname{}. Fig. \ref{fig:methodfigure} provides an overview of the model and planning procedure.

\subsection{Dataset Generation}

\begin{wrapfigure}{r}{0.45\textwidth}
\begin{minipage}{0.45\textwidth}
\vspace{-1em}
    \centering
    \includegraphics[width=\textwidth]{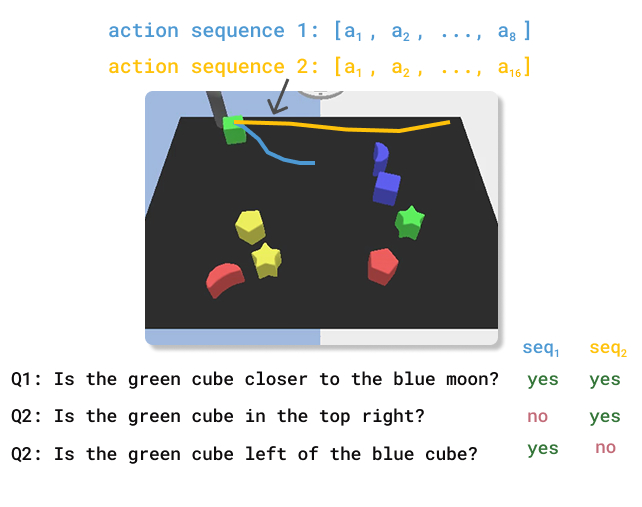}
    \vspace{-2em}
    \caption{\footnotesize{Example state entry in the SAQA dataset with two action horizons and six QA pairs.}}
    \label{fig:dataset_example}
\vspace{-2em}
\end{minipage}
\end{wrapfigure}

To train a world model to answer questions about the future, a state-action-question-answer (SAQA) dataset is generated. It is defined as
$$\mathcal{D}_{\text{SAQA}} = \{(S_i, a_{i:j}, Q_{S_j}, A_{S_j}), \dots\} \quad \text{where } j = i + h$$
where $S_i$ represents the current state (RGB frame in our case), $h$ is the horizon, $a_{i:j}$ is a sequence of actions taken from state $S_i$, and $Q_{S_j}, A_{S_j}$ is a question answer tuple about the future state $S_j$ which is reached by taking actions $a_{i:j}$ from state $S_i$. Fig. \ref{fig:dataset_example} illustrates a single state paired with multiple questions and answers in the dataset. 

The SAQA dataset is generated from a dataset of trajectories $\{T_1, T_2, \dots\}$, where each trajectory is given by a sequence of state-action tuples $\{(S_0, a_0), (S_1, a_1), \dots\}$. Here, each state comprises an image observation and privileged information, such as object positions, which are used for programmatic question generation. For each state $S_i$ in the trajectory, multiple different action horizons $h$ are sampled. As shown in Fig. \ref{fig:dataset_example}, for each sampled horizon $h$, the oracle information from future state $S_{i+h}$ is used to create a set of questions and answers, which gives the final dataset to train the model.
For each type of question generation, multiple phrasings are included in the training dataset. Examples of training question types and rewards for each task are provided in the Appendix \ref{appendix:qaprompts}. While this generation procedure currently uses state information to construct SAQA tuples, we provide preliminary evidence that off-the-shelf frontier VLMs can supply this supervision directly from images, achieving 87--92\% accuracy against ground truth across LangTable, OGBench, and DROID. This suggests a path toward removing the oracle dependence and scaling SAQA generation to real-world data; full results are reported in Appendix~\ref{appendix:autoqa}.

\subsection{Semantic World Models Architecture}
This section presents a model capable of answering questions about future events conditioned on actions. A model with such capability is fundamentally a visual question-answering model with action conditioning. Therefore, it is natural to bootstrap from large pretrained VLMs to transfer their generalization capabilities to robotics tasks. This \algname{} architecture is based on an open-source VLM, PaliGemma \citep{paligemma}. 

The model contains three core pretrained components: a transformer-based autoregressive language model with a token embedding size $d_{\text{tok}}$, a vision encoder $v_\phi$ with a feature size $d_{\text{img}}$, and a projection matrix $W \in \mathbb{R}^{d_{\text{tok}} \times d_{\text{img}}}$. The PaliGemma architecture is built on top of two individually trained components: the Gemma LLM \citep{gemmateam2024gemmaopenmodelsbased} and the SigLIP image encoder $V_{\text{sc}}$ \citep{Zhai_2023_ICCV}.  $W$ is used to project from $Z_{\text{sc}}$ to $Z_{\text{LLM}}$, where $Z_{\text{sc}}$ is the feature space of $v_\phi$, and $Z_{\text{LLM}}$ is the input token embedding space of the LLM. This paper uses the 3B parameter checkpoint from PaliGemma as the base model. This architecture and components are described in Appendix \ref{appendix:model}.

To adapt the base model to answer questions about a specific future as a result of the actions, the model needs to be conditioned on these actions. Thus, a new projection matrix $P \in \mathbb{R}^{d_{\text{tok}} \times d_{\text{act}}}$ is used which projects a single action $a \in \mathbb{R}^{d_{\text{act}}}$ into the latent space $Z_{LLM}$ similar to the $W$ projection matrix.
Given a tuple $(S_i, a_{i:j}, Q_{S_j}, A_{S_j})$ from the dataset $ \mathcal{D}_{\text{SAQA}}$, the input sequence is constructed by concatenating the image embeddings, action embeddings, and question token embeddings as $\texttt{concat}\left( W^\top V_{sc}(S_i), P^\top a_i, P^\top a_{i+1}, \dots, P^\top a_{j}, Q_{S_{j}}\right).$ The model is then fine-tuned in an end-to-end manner to predict the target answer $A_{S_{j}}$ by optimizing the standard cross-entropy loss $$\mathcal{L} = -\log p(A_{S_j} | S_i, a_{i:j}, Q_{S_j}).$$ This training procedure enables the model to capture the dynamics of the environment in language space to answer questions about future states without explicitly generating pixel-level representations.

\subsection{Planning with Semantic World Models}

Planning with world models requires evaluating the value of action sequences. For each task, a set of questions (e.g., ``is the gripper touching the block'') and desired answers (e.g., ``yes'') can be defined. For our main results, each task has a predefined set of questions and weights. We also provide ablations on the importance of these weights and on automatic question decomposition in Appendix. \ref{appendix:ablations}.
A scalar score is then derived by combining the likelihood of the model generating the desired answer for each question, weighted by some heuristic weights. Specifically, each task is defined as a set of questions, answers, and weights $\mathcal T := \{(Q_i, A_i^*, W_i)\}_{i=1}^k$. Given an observation $S$ and a sequence of actions $a_{1:n}$, its value under the task is calculated as:
\vspace{-1em}

\begin{equation}
V^{\mathcal T}(S, a_{1:n}) = \sum_{i=1}^{k} W_i \cdot p_{\text{wm}}(A_i^* | S, a_{1:n}, Q_i)
\label{eqn:value-vanilla}
\end{equation}
\vspace{-1em}

Empirical evaluation shows that rewarding the model for achieving the desired outcome earlier in the action sequence leads to better performance.
This early reward is provided by breaking each full action sequence down to sub-chunks of length $c$, and then querying the model on action sequences with increasing numbers of concatenated sub-chunks: 
\vspace{-1em}
\begin{equation}
V^{\mathcal T, c}(S, a_{1:n}) = \sum_{i=1}^{k} \sum_{\substack{j=c \\ j += c}}^{n} W_i \cdot p_{\text{wm}}(A_i^* | S, a_{1:j}, Q_i)
\label{eqn:value-multistep}
\end{equation}
\vspace{-1em}

Setting $c=1$ is equivalent to evaluating the model once for every single action in the sequence, and setting $c=h$ is equivalent to the vanilla formulation in Eqn. \ref{eqn:value-vanilla}.
Various planning techniques can be used to extract optimal actions by using the model with a well-defined value function. 
\subsubsection{Sampling-Based Planning}
Sampling-based planning provides a straightforward approach to planning with the model.
An example is Model Predictive Path Integral (MPPI) control algorithm~\cite{MPPI}, which maintains a Gaussian distribution of action parameters and iteratively refines it by querying the model. The action distribution is initialized as $\mathbf{a}^{(0)} \sim \text{Unif}(a_{\min}, a_{\max})$. At each iteration, a set of $K$ control sequences $\{ \mathbf{a}^{(k)} \}_{k=1}^K$ is sampled from the current action distribution. The value of each of these sampled trajectories $V_k$ is computed using our \algname. The distribution for the next iteration is $\mathbf{a}_{t+1} \sim \mathcal{N}\left( \mu_t, \sigma_t^2 \right)$ where
\begin{equation*}
\begin{aligned}
\mu_t = \sum_{k=1}^K  \frac{\exp\left( \frac{V_k}{\lambda} \right)}{\sum_{j=1}^K \exp\left( \frac{V_j}{\lambda} \right)} \,  \mathbf{a}_t^{(k)},  & \sigma_t^2 = \sum_{k=1}^K \omega_k \left( \mathbf{a}_t^{(k)} - \mu_t \right)^2
\end{aligned}
\end{equation*}
and $\lambda$ is a temperature parameter that controls exploration.  For our rollouts, we execute the mean
sequence of the last iteration.

\subsubsection{Gradient-Based Planning}
\label{method:gradplan}
For more complicated tasks, sampling-based planning methods typically require a large number of samples and optimization iterations, which become increasingly hard to scale for a large model like \algname.
To reduce the number of samples and model forward passes, we propose a gradient-based optimization procedure together with a base proposal policy. The gradient provides directed information for optimizing the action sequence, thus converging faster than sampling-based techniques. The base proposal policy can effectively trim down the planning search space. Given a base policy $\pi_b$, a control sequence $\mathbf{a}\sim \pi_b (S)$, and the semantic world model $p_{\text{wm}}$, gradient ascent is used to optimize the following objective:
\begin{equation}
    J^{\mathcal T}(\mathbf{a}) = V^{\mathcal T, c}(S, \textbf{a})  
\end{equation}
Where $\mathbf{a}$ is the control sequence being optimized, $\mathcal T = \{(Q_i, A_i^*, W_i)\}_{i=1}^k$ is the list of questions, desired answers, and weights, $c$ is the reward subchunk size, and $S$ is our state.
To improve stability during learning, gradient norm clipping is used before each step. Refer to Appendix \ref{appendix:visgradplan} for a visualization of this optimization process and Appendix \ref{appendix:planningspeed} to compare the computational speed of planning times for each method.

\subsection{Multistep Tasks}
To solve long-horizon tasks, the aforementioned planning procedure can be extended to a multi-step formulation. 
The capabilities of \algname{} can be used to track task progress and transition between subgoals without requiring any additional components. A series of sequential subgoals ${g_1, g_2, \dots, g_T}$ is defined where each subgoal $g_t$ is associated with a question and a desired answer corresponding to when the subgoal was completed. Each subgoal is executed sequentially and its completion is verified using \algname{}. This verification is feasible at no additional cost because zero-horizon examples are included in the training dataset. For example, in the block picking task, the following sub-goals are used: ["Is the block grasped?", "Is the block stacked on top of the other block?"],  with the desired answers ["yes", "yes"] in order to accomplish a two-stage task. This method is used to extend planning to multi-step LangTable tasks.  
\begin{figure*}[t]
    \vspace{-1em}
    \centering
    \includegraphics[width=\textwidth]{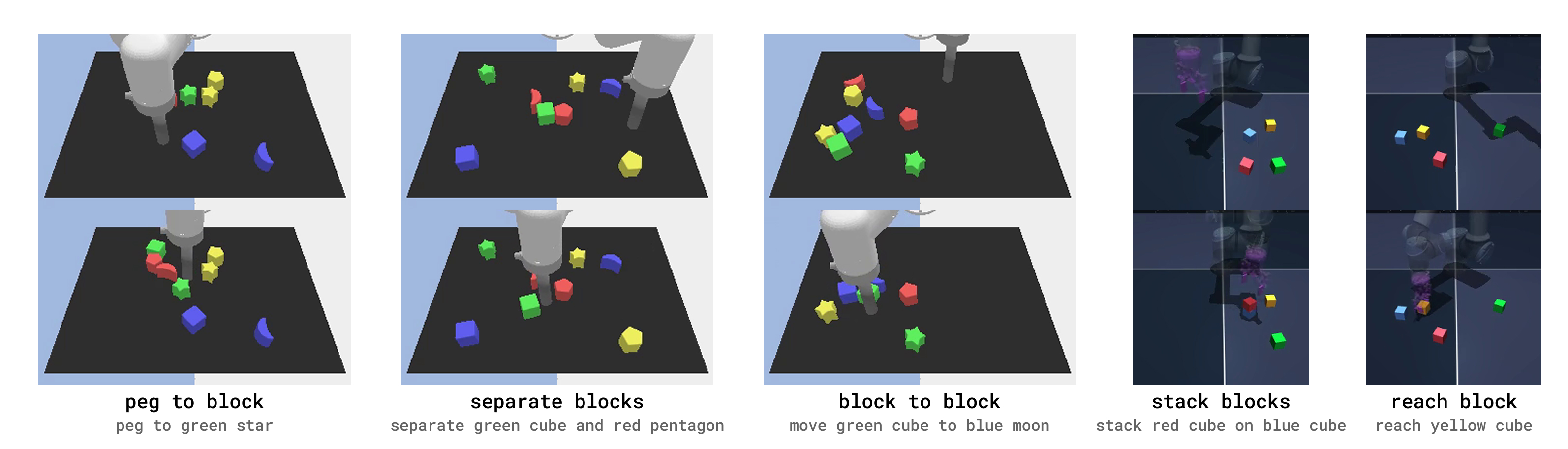}
    \caption{\footnotesize{Examples of each evaluation task. The top frame represents the initialization, and the bottom frame represents task completion. The first three tasks are for LangTable and the last two are for OGBench.}}
    \label{fig:envfigure}
\vspace{-1em}
\end{figure*}

\section{Experiments and Results}
\label{sec:experiments}

\subsection{Experimental Setup}

\algname{} is evaluated in two simulation environments, LangTable \citep{langtable} and OGBench \citep{ogbench_park2025}, capturing combinatorial generalization and dexterous manipulation. Fig. \ref{fig:envfigure} shows examples of tasks in each domain. This section provides an overview of the experiment setup and details are provided in \Cref{appendix:baselines}.

 \paragraph{LangTable} \citep{langtable} \algname{} is evaluated on \textit{reaching}, \textit{separating blocks}, and \textit{pushing} in the LangTable environment, using both sampling-based planning and gradient-based improvement over a base policy. \method is trained on a mixture of expert data collected with a scripted policy and suboptimal data collected with a random policy. To evaluate in out-of-distribution conditions, the block color combinations are changed during evaluation to test compositional generalization. For example, our training data only includes the red pentagon, and evaluation is performed with a green pentagon and a novel purple pentagon. 
\vspace{-.5em}

 \paragraph{OGBench} \citep{ogbench_park2025} In OGBench \method is evaluated on \textit{cube reaching} and a custom \textit{cube stacking} task. It is trained on a mixture of optimal and suboptimal data, collected using the provided noisy expert data and play data from OGBench, respectively. Background color is changed during evaluation to measure generalization.

For both environments, a per-task Diffusion Policy \citep{diffpol} is trained on 300 expert trajectories for 100 epochs as the base policy. The expert trajectories were collected using the same experts as in the offline dataset.

During training, the dataset was balanced in both the number of each possible question type and the answer distribution for each respective question. For example, for each state in the LangTable environment, there are $\binom{8}{2}$ possible questions about whether two blocks are touching, but $8$ questions about whether the end effector is touching a given block. Similarly, most blocks are separated in the initial states of the LangTable environment, leading to far more `no' answers than `yes' answers. The imbalance is addressed during training by oversampling tuples such that there is a balanced amount of question types and answer distributions.
\vspace{-.5em}

\begin{figure}[t]
    \centering
    \includegraphics[width=.8\linewidth]{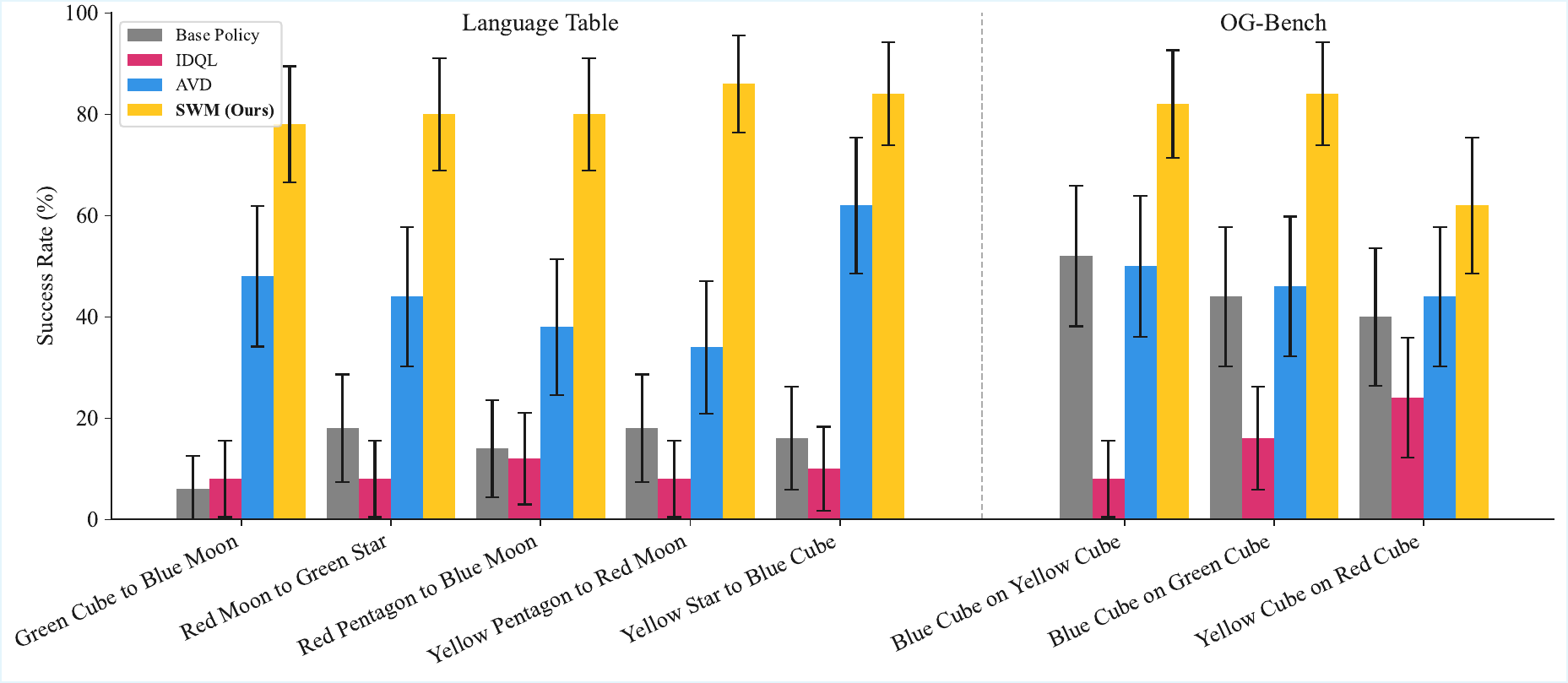}
    \caption{\footnotesize{\textbf{Policy Improvement} across LangTable and OGBench across multiple tasks. The average success rates of the base policies ($14.4\%$ on LangTable and $45.33\%$ on OGBench) increase to $81.6\%$ and $76.0\%$, respectively. SWM further outperforms the IDQL and AVD baselines across all evaluated tasks and environments.  Reported success rates over $n=50$ seeds with 95\% confidence intervals (normal approximation). }}
    \label{fig:improvement}
    \vspace{-1em}

\end{figure}
\subsection{Baselines}
\methodfulls is compared to the baselines below. Details about each baseline and hyperparameters are described in Appendix \ref{appendix:baselines}.

 \textbf{IDQL} \citep{idql}: IDQL is an offline RL baseline which uses IQL \cite{kostrikov2022offline} to reweight a behavior diffusion policy. For each task, the offline dataset used for \methodfull is combined with the per-task expert dataset used for the base policy. This combined dataset is labeled with binary rewards and used to train the IDQL policy. The architecture and hyperparameters of the diffusion policy used as the IDQL behavior policy are the same as for the base policies, except with a horizon of 8.

\textbf{Action Conditioned Video Diffusion (AVD)}: To isolate the benefits of modeling only task-relevant semantics, we introduce a pixel-based world model baseline. We train a diffusion transformer model to predict future observations conditioned on actions. In order to obtain learning signal from these raw pixel predictions, this baseline first generates the future observation and then extracts the necessary semantic information using VQA. The resulting outputs are used as rewards for MPPI planning. This baseline therefore represents a conventional pixel-centric approach: extracting task-relevant information from the generated visual observations, rather than modeling semantic information directly.  The initial trajectory candidate samples are generated through the same base diffusion policy used in SWM. To ensure a fair comparison, we prioritize AVD's generation quality (using 100 diffusion steps) over inference speed.

\textbf{Latent World Models}: We additionally evaluated two latent world modeling approaches, Dreamer \citep{Hafner2020Dream} and TDMPC2 \citep{hansen2024tdmpc}, online model based RL approaches, which we modify to use our offline dataset and sparse rewards. We found that neither was able to successfully solve the LangTable tasks under our setting, so we omit their performance from the main results. Additional details are provided in Appendix~\ref{appendix:dreamer}.

\subsection{Results}

The evaluation aims to address the following questions: (1) Is \method an effective world model for decision making? (2) Does suboptimal data improve modeling performance? (3) Does \method preserve the generalization capabilities from the base VLM?
\begin{table}[t]
    \centering
    \resizebox{\textwidth}{!}{
    \begin{tabular}{lcccc}
    \toprule
         & \multicolumn{2}{c}{\small \textit{LangTable}} & \multicolumn{2}{c}{\small \textit{OGBench}} \\

    Dataset Type \hspace{2cm} & Expert Data & Expert Data OOD & Expert Data & Expert Data OOD\\
    \midrule

    Sub Optimal & \(85.98 \pm 0.33\) & \(81.99 \pm 1.46\) & \(90.83 \pm 0.39\) & \(85.56 \pm 1.10\) \\

    Expert & \(91.27 \pm 0.79\) & \(86.49 \pm 0.39\) & \(96.53 \pm 0.13\) & \(87.33 \pm 2.13\) \\

    Combined & \(\textbf{92.92} \pm 0.34\) & \(\textbf{88.32} \pm 2.10\) & \(\textbf{96.86} \pm 0.13\) & \(\textbf{88.16} \pm 1.54\) \\
    \bottomrule
    
    \end{tabular}
    }
    \caption{\footnotesize{\textbf{Future QA Performance.} Accuracy of answers on future QA evaluated on expert SAQA datasets generated by experts on test time seeds in both in-domain and out-of-domain block combinations. Reported standard deviation across 3 model training seeds.}}
    
    \label{tab:future_qa}
\end{table}

\paragraph{Is \method an effective world model for decision making? } 

\begin{table}[b]
    \centering
    \small

    \begin{tabular}{lccc}
    \toprule
    Task & Base Policy & AVD & SWM (Ours) \\
    \midrule
    MS1 & 6\% $\pm$ 6.6 & 8\% $\pm$ 7.5 & \textbf{50\%} $\pm$ 13.9\\
    MS2 & 4\% $\pm$ 5.4 & 2\% $\pm$ 3.9 & \textbf{66\%} $\pm$ 13.1\\
    MS3 & 4\% $\pm$ 5.4 & 2\% $\pm$ 3.9 & \textbf{54\%} $\pm$ 13.8\\
    MS4 & 2\% $\pm$ 3.9 & 4\% $\pm$ 5.4 & \textbf{54\%} $\pm$ 13.8\\
    \bottomrule
    \end{tabular}

       \caption{\footnotesize{\textbf{Multi-Step Results.} SWM model improvement results on four different multi-step compositional tasks. The tasks are as follows: MS1 - red pentagon to blue moon, yellow pentagon to red moon. MS2 - yellow star to blue cube, yellow pentagon to red moon. MS3 - yellow star to blue cube, red pentagon to blue moon. MS4 - green cube to blue moon, yellow pentagon to red moon. Reported success rates over $n=50$ seeds with 95\% confidence intervals (normal approximation).}}
    \label{tab:multistep}
    \vspace{-2em}
\end{table}

\vspace{-.5em}
The planning capabilities of SWM is evaluated first by applying a sampling-based planning method, MPPI, to a \method model on LangTable and OGBench tasks.
As shown in Tab. \ref{tab:planning_results}, it is possible to directly plan on top of the semantic world model using sampling-based planning methods, achieving close to perfect success rates on reaching and block separation tasks across both environments. 

However, the computational cost of the sampling-based planning method with large models makes it infeasible to run MPPI on more challenging tasks requiring a higher number of samples.  Therefore, for more complicated tasks, consider a scenario in which a base policy generates a candidate trajectory that is refined using \method and gradient-based optimization (described in \Cref{method:gradplan}).  As shown in Fig. \ref{fig:improvement}, the method is able to refine candidate trajectories and show substantial improvement over the base policies. SWM demonstrates an average performance increase over the base policies from $14.4\%$ to $81.6\%$ on average for LangTable and $45.33\%$ to $76\%$ on average for OGBench.  \algname{} also outperforms both the AVD and IDQL baselines across all tasks, demonstrating the effectiveness of \algname{} for planning. 
\begin{wraptable}{r}{0.35\textwidth}
\begin{minipage}{0.35\textwidth}

    \vspace{-.5em}
    \centering
    \begin{tabular}{lccc}
        \toprule
        Task & SWM \\
        \midrule  
        
        LT Reach Block & 100\% \\
    
        LT Separate Blocks & 100\% \\
        \midrule
        OG Reach Cube &  97\%\\
    
        \bottomrule
    \end{tabular}
    \caption{\footnotesize{\textbf{Planning Results} MPPI planning success rates over 100 seeds.}}
    \label{tab:planning_results}
    \vspace{-2em}
\end{minipage}
\end{wraptable}

\algname{} also demonstrates the capability for longer horizon tasks by both selecting subgoals and then planning using that specific subgoal. SWM demonstrates an average policy improvement of $52.0\%$ as shown in Tab. \ref{tab:multistep} on multistep tasks, outperforming the AVD baseline. For both AVD and \algname{}, subgoal completion was determined using the SWM model without action conditioning.

\paragraph{Does suboptimal data improve modeling performance?} One of the key aspects of a world model is its ability to learn from suboptimal data. To measure the effects of suboptimal demonstrations, a test set of future QA data collected from expert demonstrations in both the in-distribution and out-of-distribution environments is created. The models are then trained on three different seeds and fixed hyperparameters to convergence with suboptimal data, optimal data, or a 50/50 mixed dataset. As seen in Table~\ref{tab:future_qa}, mixing in the suboptimal data improves accuracy over training on just expert data. \method is also able to achieve moderate levels of performance by training only on suboptimal data, demonstrating how effective suboptimal data can be for training our world model.

\paragraph{Does training preserve the generalization capabilities from the base VLM?} To measure the effects of VLM pretraining on generalization, \method is evaluated on compositional and scene out-of-distribution environments, depicted in Fig. \ref{fig:ood_figure}. Since the offline dataset was misaligned with these evaluation tasks, the IDQL baseline is not evaluated.

\begin{table}[t]
    \vspace{-1em}
    \centering
    \small

    \begin{tabular}{lccc}
    \toprule
    Task & Base Policy & AVD & SWM (Ours) \\
    \midrule
    Push Blue Star to Red Cube & 54\% $\pm$ 13.8 & 66\% $\pm$ 13.1 & \textbf{86\%} $\pm$ 9.6\\
    Push Yellow Moon to Purple Cube & 54\% $\pm$ 13.8 & 56\% $\pm$ 13.8 & \textbf{78\%} $\pm$ 11.5\\
    Stack Red to Green OOD Background & 62\% $\pm$ 13.5 & 28\% $\pm$ 12.4 & \textbf{72\%} $\pm$ 12.4\\
    Stack Blue to Yellow OOD Background & 50\% $\pm$ 13.9 & 50\% $\pm$ 13.9 & \textbf{70\%} $\pm$ 12.7\\
    \bottomrule
    \end{tabular}    
    \caption{\footnotesize{\textbf{Out-of-Distribution Improvement Results.} SWM model improvement results on tasks in LangTable and OGBench on out-of-distribution scenes. \algname{} is able to show policy improvement and outperform AVD across both environments. Reported success rates over $n=50$ seeds with 95\% confidence intervals (normal approximation).}}
    \label{tab:improvementOOD}
    \vspace{-.5em}

\end{table}

To measure semantic compositional generalization, a new colored block is introduced and the existing block color-shape pairs are modified in the LangTable environment. Tab. \ref{tab:improvementOOD} shows an average improvement of $28.0\%$ percentage points over the base policies under these conditions. This performance indicates that \method is able to retain some of the pretraining knowledge, resulting in compositional generalization. 

To test robustness to background changes, OGBench's background color is changed to a novel combination. \method is again able to demonstrate a $15\%$ precentage points boost in performance compared to the base policy and is able to generalize to these conditions, while the AVD method is unable to. 

\begin{figure*}[t]
    \centering
    \includegraphics[width=.7\textwidth]{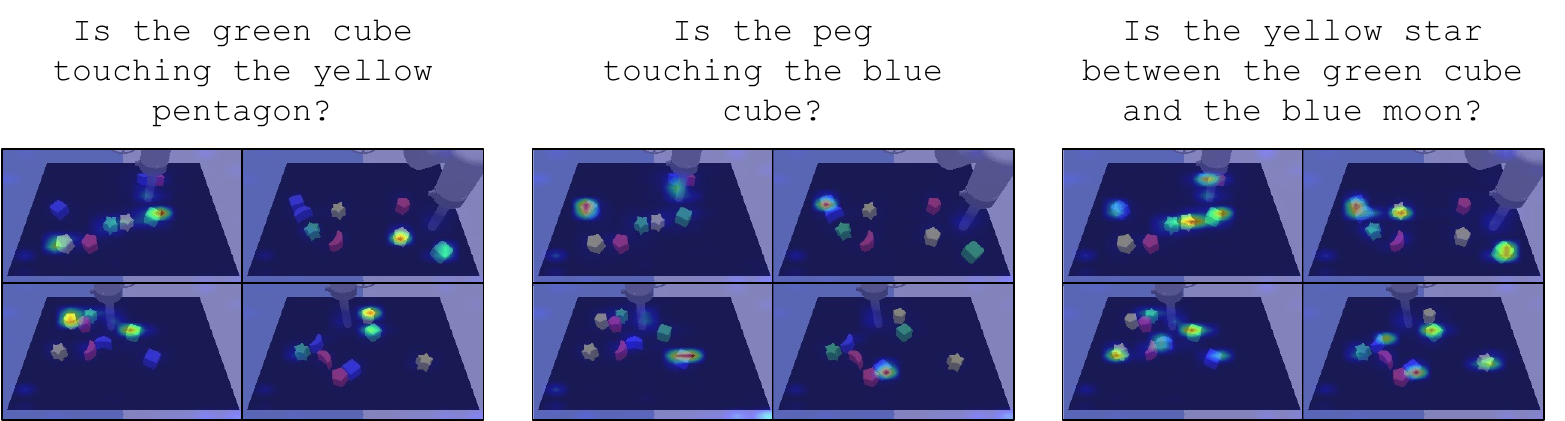}
    \caption{\footnotesize{Visualization of the attention map from language tokens to image patches in the 4th transformer layer. The language tokens correctly attend to the task-relevant locations in the image depending on the prompt.}}
    \label{fig:attn}
    \vspace{-1.5em}
\end{figure*}

\paragraph{Does the model's internal representations attend to the task-relevant information?}
To understand the learned representations of the model, the attention maps from the language tokens to the image patches are visualized from an intermediate layer of the model. As shown in Fig. \ref{fig:attn}, the model correctly attends to the task-relevant location in the image depending on the language prompt. For example, when asked ``Is the red moon touching the blue cube?", the attention score is higher on the image patches corresponding to the objects. Although never finetuned on questions with more than two objects, the model was found to correctly attend to three objects when asked to. This shows that the model inherits generalization from the pretrained VLM. In Appendix \ref{appendix:attn} more visualizations of individual layers as well as entire trajectories are provided.

\section{Conclusions}

This paper presents semantic world models, a novel world modeling approach that explicitly models future outcomes through QA about the consequences without needing to reconstruct or use pixel-level information as a training objective. It shows that this approach can be used with both sampling-based planning and gradient-based policy improvement. Empirical evaluation demonstrates considerable gains over pixel and latent based world modeling and offline RL methods, suggesting \method could be the basis of a new framework for world modeling. These benchmarks were selected to highlight complementary aspects of the approach, covering long-horizon and compositional tasks (LangTable), and tabletop manipulation (OGBench), providing meaningful evidence for the viability of the SWM paradigm.

\paragraph{Limitations and Future Work.}
While Semantic World Models demonstrate strong performance on these tasks, several limitations remain that suggest directions for extending this paradigm to real-world settings. First, the high parameter count of the base VLM makes sample-based planning methods computationally prohibitive to perform on a single GPU or at a reasonable control frequency. Although gradient-based planning is substantially more efficient, it depends on a base policy to generate an initial trajectory. Recent advances in compact VLMs, such as FastVLM and SmolVLM \citep{smolvlm, fastvlm}, suggest a promising path toward scalable sample-based planning without reliance on a base policy. Second, \method currently uses ground truth state information to construct the SAQA dataset, which is difficult to obtain in real-world environments. Replacing oracle-generated QA pairs with those bootstrapped from a base VLM could significantly increase data diversity and enable the incorporation of real-world data into the training of Semantic World Models. We present preliminary results evaluating the accuracy and forward progress of frontier VLMs for this form of spatial reasoning supervision in Appendix~\ref{appendix:autoqa}. Finally, the current formulation uses binary yes/no questions for the world model, although the framework should conceptually cover richer QA pairs; extending the QA framework to richer, open-ended responses or including reasoning could unlock more expressive semantic supervision.

\subsubsection*{Acknowledgments}
This material is based upon work supported by the National Science Foundation under Grant No. 2212310. This work was also supported through funding from the Army Research Lab.   

\bibliography{iclr2026_conference}

@misc{langtable,
      title={Interactive Language: Talking to Robots in Real Time}, 
      author={Corey Lynch and Ayzaan Wahid and Jonathan Tompson and Tianli Ding and James Betker and Robert Baruch and Travis Armstrong and Pete Florence},
      year={2022},
      eprint={2210.06407},
      archivePrefix={arXiv},
      primaryClass={cs.RO},
      url={https://arxiv.org/abs/2210.06407}, 
}

@misc{paligemma,
      title={PaliGemma: A versatile 3B VLM for transfer}, 
      author={Lucas Beyer and Andreas Steiner and André Susano Pinto and Alexander Kolesnikov and Xiao Wang and Daniel Salz and Maxim Neumann and Ibrahim Alabdulmohsin and Michael Tschannen and Emanuele Bugliarello and Thomas Unterthiner and Daniel Keysers and Skanda Koppula and Fangyu Liu and Adam Grycner and Alexey Gritsenko and Neil Houlsby and Manoj Kumar and Keran Rong and Julian Eisenschlos and Rishabh Kabra and Matthias Bauer and Matko Bošnjak and Xi Chen and Matthias Minderer and Paul Voigtlaender and Ioana Bica and Ivana Balazevic and Joan Puigcerver and Pinelopi Papalampidi and Olivier Henaff and Xi Xiong and Radu Soricut and Jeremiah Harmsen and Xiaohua Zhai},
      year={2024},
      eprint={2407.07726},
      archivePrefix={arXiv},
      primaryClass={cs.CV},
      url={https://arxiv.org/abs/2407.07726}, 
}

@inproceedings{diffpol,
	title={Diffusion Policy: Visuomotor Policy Learning via Action Diffusion},
	author={Chi, Cheng and Feng, Siyuan and Du, Yilun and Xu, Zhenjia and Cousineau, Eric and Burchfiel, Benjamin and Song, Shuran},
	booktitle={Proceedings of Robotics: Science and Systems (RSS)},
	year={2023}
}

@misc{idql,
      title={IDQL: Implicit Q-Learning as an Actor-Critic Method with Diffusion Policies}, 
      author={Philippe Hansen-Estruch and Ilya Kostrikov and Michael Janner and Jakub Grudzien Kuba and Sergey Levine},
      year={2023},
      eprint={2304.10573},
      archivePrefix={arXiv},
      primaryClass={cs.LG},
      url={https://arxiv.org/abs/2304.10573}, 
}

@misc{smolvlm,
      title={SmolVLM: Redefining small and efficient multimodal models}, 
      author={Andrés Marafioti and Orr Zohar and Miquel Farré and Merve Noyan and Elie Bakouch and Pedro Cuenca and Cyril Zakka and Loubna Ben Allal and Anton Lozhkov and Nouamane Tazi and Vaibhav Srivastav and Joshua Lochner and Hugo Larcher and Mathieu Morlon and Lewis Tunstall and Leandro von Werra and Thomas Wolf},
      year={2025},
      eprint={2504.05299},
      archivePrefix={arXiv},
      primaryClass={cs.AI},
      url={https://arxiv.org/abs/2504.05299}, 
}

@misc{fastvlm,
  author = {Pavan Kumar and Anasosalu Vasu and Fartash Faghri and Chun-Liang Li and Cem Koc and Nate True and Albert Antony and Gokul Santhanam and James Gabriel and Peter Grasch and Oncel Tuzel and Hadi Pouransari},
  title = {FastVLM: Efficient Vision Encoding for Vision Language Models},
  month = {June},
  year = {2025},
}

@misc{genie3,
  title         = {Genie 3: A New Frontier for World Models},
  author        = {Philip J. Ball and Jakob Bauer and Frank Belletti and Bethanie Brownfield and Ariel Ephrat and Shlomi Fruchter and Agrim Gupta and Kristian Holsheimer and Aleksander Holynski and Jiri Hron and Christos Kaplanis and Marjorie Limont and Matt McGill and Yanko Oliveira and Jack Parker-Holder and Frank Perbet and Guy Scully and Jeremy Shar and Stephen Spencer and Omer Tov and Ruben Villegas and Emma Wang and Jessica Yung and Cip Baetu and Jordi Berbel and David Bridson and Jake Bruce and Gavin Buttimore and Sarah Chakera and Bilva Chandra and Paul Collins and Alex Cullum and Bogdan Damoc and Vibha Dasagi and Maxime Gazeau and Charles Gbadamosi and Woohyun Han and Ed Hirst and Ashyana Kachra and Lucie Kerley and Kristian Kjems and Eva Knoepfel and Vika Koriakin and Jessica Lo and Cong Lu and Zeb Mehring and Alex Moufarek and Henna Nandwani and Valeria Oliveira and Fabio Pardo and Jane Park and Andrew Pierson and Ben Poole and Helen Ran and Tim Salimans and Manuel Sanchez and Igor Saprykin and Amy Shen and Sailesh Sidhwani and Duncan Smith and Joe Stanton and Hamish Tomlinson and Dimple Vijaykumar and Luyu Wang and Piers Wingfield and Nat Wong and Keyang Xu and Christopher Yew and Nick Young and Vadim Zubov and Douglas Eck and Dumitru Erhan and Koray Kavukcuoglu and Demis Hassabis and Zoubin Gharamani and Raia Hadsell and A{\"a}ron van den Oord and Inbar Mosseri and Adrian Bolton and Satinder Singh and Tim Rockt{\"a}schel},
  year          = {2025},
  url           = {}
}

@misc{hansen2022temporaldifferencelearningmodel,
      title={Temporal Difference Learning for Model Predictive Control}, 
      author={Nicklas Hansen and Xiaolong Wang and Hao Su},
      year={2022},
      eprint={2203.04955},
      archivePrefix={arXiv},
      primaryClass={cs.LG},
      url={https://arxiv.org/abs/2203.04955}, 
}

@INPROCEEDINGS{MPPI,
  author={Williams, Grady and Drews, Paul and Goldfain, Brian and Rehg, James M. and Theodorou, Evangelos A.},
  booktitle={2016 IEEE International Conference on Robotics and Automation (ICRA)}, 
  title={Aggressive driving with model predictive path integral control}, 
  year={2016},
  volume={},
  number={},
  pages={1433-1440},
  doi={10.1109/ICRA.2016.7487277}}

@InProceedings{radford21clip,
  title = 	 {Learning Transferable Visual Models From Natural Language Supervision},
  author =       {Radford, Alec and Kim, Jong Wook and Hallacy, Chris and Ramesh, Aditya and Goh, Gabriel and Agarwal, Sandhini and Sastry, Girish and Askell, Amanda and Mishkin, Pamela and Clark, Jack and Krueger, Gretchen and Sutskever, Ilya},
  booktitle = 	 {Proceedings of the 38th International Conference on Machine Learning},
  pages = 	 {8748--8763},
  year = 	 {2021},
  editor = 	 {Meila, Marina and Zhang, Tong},
  volume = 	 {139},
  series = 	 {Proceedings of Machine Learning Research},
  month = 	 {18--24 Jul},
  publisher =    {PMLR},
  url = 	 {https://proceedings.mlr.press/v139/radford21a.html}
}

@InProceedings{Zhai_2023_ICCV,
  author    = {Zhai, Xiaohua and Mustafa, Basil and Kolesnikov, Alexander and Beyer, Lucas},
  title     = {Sigmoid Loss for Language Image Pre-Training},
  booktitle = {Proceedings of the IEEE/CVF International Conference on Computer Vision (ICCV)},
  year      = {2023},
  pages     = {11975--11986},
  month     = {October}
}

@misc{gemini2023,
  title        = {Gemini: A Family of Highly Capable Multimodal Models},
  author       = {{Gemini Team}},
  year         = {2023},
  eprint       = {2312.11805},
  archivePrefix= {arXiv},
  primaryClass = {cs.CL},
  url          = {https://arxiv.org/abs/2312.11805}
}

@article{Molmo_PixMo_2024,
  title        = {Molmo and PixMo: Open Weights and Open Data for State-of-the‐Art Multimodal Models},
  author       = {M. Deitke and others},
  journal      = {arXiv preprint arXiv:2409.17146},
  year         = {2024},
  eprint       = {2409.17146},
  archivePrefix= {arXiv},
  url          = {https://arxiv.org/abs/2409.17146}
}

@article{Qwen_2023,
  title        = {Qwen Technical Report},
  author       = {Bai, J. and Bai, S. and Yang, S. and others},
  journal      = {arXiv preprint arXiv:2309.16609},
  year         = {2023},
  eprint       = {2309.16609},
  archivePrefix= {arXiv},
  url          = {https://arxiv.org/abs/2309.16609}
}

@article{LLaMA_2023,
  title        = {LLaMA: Open and Efficient Foundation Language Models},
  author       = {Touvron, Hugo and Lavril, Thibaut and Izacard, Gautier and Martinet, Xavier and Lachaux, Marie-Anne and Lacroix, Timothée and Rozière, Baptiste and Goyal, Naman and Hambro, Eric and Azhar, Faisal and Rodriguez, Aurélien and Joulin, Armand and Grave, Edouard and Lample, Guillaume},
  journal      = {CoRR},
  volume       = {abs/2302.13971},
  year         = {2023},
  eprint       = {2302.13971},
  archivePrefix= {arXiv},
  doi          = {10.48550/arXiv.2302.13971},
  url          = {https://arxiv.org/abs/2302.13971}
}

@misc{GPT4o_2024,
  title        = {GPT-4o System Card},
  author       = {OpenAI},
  year         = {2024},
  eprint       = {2410.21276},
  archivePrefix= {arXiv},
  url          = {https://arxiv.org/abs/2410.21276}
}

@InProceedings{BrohanBrownCarbajalEtAl_2023_RT1,
  author       = {Brohan, Anthony and Brown, Noah and Carbajal, Justice and Chebotar, Yevgen and Dabis, Joseph and Finn, Chelsea and Gopalakrishnan, Keerthana and Hausman, Karol and Herzog, Alexander and Hsu, Jasmine and Ibarz, Julian and Ichter, Brian and Irpan, Alex and Jackson, Tomas and Jesmonth, Sally and Joshi, Nikhil J. and Julian, Ryan and Kalashnikov, Dmitry and Kuang, Yuheng and Leal, Isabel and Lee, Kuang-Huei and Levine, Sergey and Lu, Yao and Malla, Utsav and Manjunath, Deeksha and Mordatch, Igor and Nachum, Ofir and Parada, Carolina and Peralta, Jodilyn and Perez, Emily and Pertsch, Karl and Quiambao, Jornell and Rao, Kanishka and Ryoo, Michael and Salazar, Grecia and Sanketi, Pannag and Sayed, Kevin and Singh, Jaspiar and Sontakke, Sumedh and Stone, Austin and Tan, Clayton and Tran, Huong and Vanhoucke, Vincent and Vega, Steve and Vuong, Quan and Xia, Fei and Xiao, Ted and Xu, Peng and Xu, Sichun and Yu, Tianhe and Zitkovich, Brianna},
  title        = {RT-1: Robotics Transformer for Real-World Control at Scale},
  booktitle    = {Proceedings of Robotics: Science and Systems (RSS)},
  year         = {2023},
  month        = {July},
  address      = {Daegu, Republic of Korea},
  doi          = {10.15607/RSS.2023.XIX.025},
}

@InProceedings{KimPertschKaramchetiEtAl_2025_OpenVLA,
  author       = {Kim, Moo Jin and Pertsch, Karl and Karamcheti, Siddharth and Xiao, Ted and Balakrishna, Ashwin and Nair, Suraj and Rafailov, Rafael and Foster, Ethan P. and Sanketi, Pannag R. and Vuong, Quan and Kollar, Thomas and Burchfiel, Benjamin and Tedrake, Russ and Sadigh, Dorsa and Levine, Sergey and Liang, Percy and Finn, Chelsea},
  title        = {OpenVLA: An Open-Source Vision-Language-Action Model},
  booktitle    = {Proceedings of The 8th Conference on Robot Learning (CoRL)},
  series       = {Proceedings of Machine Learning Research},
  volume       = {270},
  pages        = {2679--2713},
  year         = {2025},
  month        = {November},
}

@Misc{BlackBrownDarpinianEtAl_2024_Pi0,
  author       = {Black, Kevin and Brown, Noah and Darpinian, James and Dhabalia, Karan and Driess, Danny and Esmail, Adnan and Equi, Michael and Fusai, Niccolo and Galliker, Manuel Y. and Ghosh, Dibya and Groom, Lachy and Hausman, Karol and Ichter, Brian and Shi, Lucy Xiaoyang and Smith, Laura and Springenberg, Jost Tobias and Stachowicz, Kyle and Tanner, James and Vuong, Quan and Walke, Homer and Walling, Anna and Wang, Haohuan and Yu, Lili and Zhilinsky, Ury},
  title        = {$\pi_0$: A Vision-Language-Action Flow Model for General Robot Control},
  year         = {2024},
  howpublished = {arXiv preprint arXiv:2410.24164},
  note         = {submitted October 2024},
}

@InProceedings{Chua2018PETS,
  title        = {Deep Reinforcement Learning in a Handful of Trials using Probabilistic Dynamics Models},
  author       = {Chua, Kurtland and Calandra, Roberto and McAllister, Rowan and Levine, Sergey},
  booktitle    = {Advances in Neural Information Processing Systems (NeurIPS)},
  volume       = {31},
  pages        = {4754--4765},
  year         = {2018}
}

@InProceedings{Hafner2019PlaNet,
  title        = {Learning Latent Dynamics for Planning from Pixels},
  author       = {Hafner, Danijar and Lillicrap, Timothy and Fischer, Ian and Villegas, Ruben and Ha, David and Lee, Honglak and Davidson, James},
  booktitle    = {Proceedings of the 36th International Conference on Machine Learning (ICML)},
  pages        = {2555--2565},
  year         = {2019}
}

@InProceedings{Rybkin2021LatCo,
  title        = {Model-Based Reinforcement Learning via Latent-Space Collocation},
  author       = {Rybkin, Oleh and Zhu, Chuning and Nagabandi, Anusha and Daniilidis, Kostas and Mordatch, Igor and Levine, Sergey},
  booktitle    = {Proceedings of the 38th International Conference on Machine Learning (ICML)},
  volume       = {139},
  pages        = {8691--8702},
  year         = {2021}
}

@inproceedings{
Hafner2020Dream,
title={Dream to Control: Learning Behaviors by Latent Imagination},
author={Danijar Hafner and Timothy Lillicrap and Jimmy Ba and Mohammad Norouzi},
booktitle={International Conference on Learning Representations},
year={2020},
url={https://openreview.net/forum?id=S1lOTC4tDS}
}

@inproceedings{zhu2025uwm,
    author    = {Zhu, Chuning and Yu, Raymond and Feng, Siyuan and Burchfiel, Benjamin and Shah, Paarth and Gupta, Abhishek},
    title     = {Unified World Models: Coupling Video and Action Diffusion for Pretraining on Large Robotic Datasets},
    booktitle = {Proceedings of Robotics: Science and Systems (RSS)},
    year      = {2025},
}

@inproceedings{
zhu2023repo,
title={RePo: Resilient Model-Based Reinforcement Learning by Regularizing Posterior Predictability},
author={Chuning Zhu and Max Simchowitz and Siri Gadipudi and Abhishek Gupta},
booktitle={Thirty-seventh Conference on Neural Information Processing Systems},
year={2023},
url={https://openreview.net/forum?id=OIJ3VXDy6s}
}

@inproceedings{
du2023learning,
title={Learning Universal Policies via Text-Guided Video Generation},
author={Yilun Du and Sherry Yang and Bo Dai and Hanjun Dai and Ofir Nachum and Joshua B. Tenenbaum and Dale Schuurmans and Pieter Abbeel},
booktitle={Thirty-seventh Conference on Neural Information Processing Systems},
year={2023},
url={https://openreview.net/forum?id=bo8q5MRcwy}
}

@inproceedings{
zhang2021learning,
title={Learning Invariant Representations for Reinforcement Learning without Reconstruction},
author={Amy Zhang and Rowan Thomas McAllister and Roberto Calandra and Yarin Gal and Sergey Levine},
booktitle={International Conference on Learning Representations},
year={2021},
url={https://openreview.net/forum?id=-2FCwDKRREu}
}

@misc{adamw,
      title={Decoupled Weight Decay Regularization}, 
      author={Ilya Loshchilov and Frank Hutter},
      year={2019},
      eprint={1711.05101},
      archivePrefix={arXiv},
      primaryClass={cs.LG},
      url={https://arxiv.org/abs/1711.05101}, 
}

@misc{gemmateam2024gemmaopenmodelsbased,
      title={Gemma: Open Models Based on Gemini Research and Technology}, 
      author={{Gemma Team} and Thomas Mesnard and Cassidy Hardin and Robert Dadashi and Surya Bhupatiraju and Shreya Pathak and Laurent Sifre and Morgane Rivière and Mihir Sanjay Kale and Juliette Love and Pouya Tafti and Léonard Hussenot and Pier Giuseppe Sessa and Aakanksha Chowdhery and Adam Roberts and Aditya Barua and Alex Botev and Alex Castro-Ros and Ambrose Slone and Amélie Héliou and Andrea Tacchetti and Anna Bulanova and Antonia Paterson and Beth Tsai and Bobak Shahriari and Charline Le Lan and Christopher A. Choquette-Choo and Clément Crepy and Daniel Cer and Daphne Ippolito and David Reid and Elena Buchatskaya and Eric Ni and Eric Noland and Geng Yan and George Tucker and George-Christian Muraru and Grigory Rozhdestvenskiy and Henryk Michalewski and Ian Tenney and Ivan Grishchenko and Jacob Austin and James Keeling and Jane Labanowski and Jean-Baptiste Lespiau and Jeff Stanway and Jenny Brennan and Jeremy Chen and Johan Ferret and Justin Chiu and Justin Mao-Jones and Katherine Lee and Kathy Yu and Katie Millican and Lars Lowe Sjoesund and Lisa Lee and Lucas Dixon and Machel Reid and Maciej Mikuła and Mateo Wirth and Michael Sharman and Nikolai Chinaev and Nithum Thain and Olivier Bachem and Oscar Chang and Oscar Wahltinez and Paige Bailey and Paul Michel and Petko Yotov and Rahma Chaabouni and Ramona Comanescu and Reena Jana and Rohan Anil and Ross McIlroy and Ruibo Liu and Ryan Mullins and Samuel L Smith and Sebastian Borgeaud and Sertan Girgin and Sholto Douglas and Shree Pandya and Siamak Shakeri and Soham De and Ted Klimenko and Tom Hennigan and Vlad Feinberg and Wojciech Stokowiec and Yu-hui Chen and Zafarali Ahmed and Zhitao Gong and Tris Warkentin and Ludovic Peran and Minh Giang and Clément Farabet and Oriol Vinyals and Jeff Dean and Koray Kavukcuoglu and Demis Hassabis and Zoubin Ghahramani and Douglas Eck and Joelle Barral and Fernando Pereira and Eli Collins and Armand Joulin and Noah Fiedel and Evan Senter and Alek Andreev and Kathleen Kenealy},
      year={2024},
      eprint={2403.08295},
      archivePrefix={arXiv},
      primaryClass={cs.CL},
      url={https://arxiv.org/abs/2403.08295}, 
}

@inproceedings{ogbench_park2025,
  title={OGBench: Benchmarking Offline Goal-Conditioned RL},
  author={Park, Seohong and Frans, Kevin and Eysenbach, Benjamin and Levine, Sergey},
  booktitle={International Conference on Learning Representations (ICLR)},
  year={2025},
}

@inproceedings{locatello2020slot,
 author = {Locatello, Francesco and Weissenborn, Dirk and Unterthiner, Thomas and Mahendran, Aravindh and Heigold, Georg and Uszkoreit, Jakob and Dosovitskiy, Alexey and Kipf, Thomas},
 booktitle = {Advances in Neural Information Processing Systems},
 editor = {H. Larochelle and M. Ranzato and R. Hadsell and M.F. Balcan and H. Lin},
 pages = {11525--11538},
 publisher = {Curran Associates, Inc.},
 title = {Object-Centric Learning with Slot Attention},
 url = {https://proceedings.neurips.cc/paper_files/paper/2020/file/8511df98c02ab60aea1b2356c013bc0f-Paper.pdf},
 volume = {33},
 year = {2020}
}

@article{
bo2025areview,
author = {Bo Ai  and Stephen Tian  and Haochen Shi  and Yixuan Wang  and Tobias Pfaff  and Cheston Tan  and Henrik I. Christensen  and Hao Su  and Jiajun Wu  and Yunzhu Li },
title = {A review of learning-based dynamics models for robotic manipulation},
journal = {Science Robotics},
volume = {10},
number = {106},
pages = {eadt1497},
year = {2025},
doi = {10.1126/scirobotics.adt1497},
URL = {https://www.science.org/doi/abs/10.1126/scirobotics.adt1497},
eprint = {https://www.science.org/doi/pdf/10.1126/scirobotics.adt1497}}

@inproceedings{
hansen2024tdmpc,
title={{TD}-{MPC}2: Scalable, Robust World Models for Continuous Control},
author={Nicklas Hansen and Hao Su and Xiaolong Wang},
booktitle={The Twelfth International Conference on Learning Representations},
year={2024},
url={https://openreview.net/forum?id=Oxh5CstDJU}
}

@book{cem,
author = {Rubinstein, Reuven Y. and Kroese, Dirk P.},
title = {The Cross Entropy Method: A Unified Approach To Combinatorial Optimization, Monte-carlo Simulation (Information Science and Statistics)},
year = {2004},
isbn = {038721240X},
publisher = {Springer-Verlag},
address = {Berlin, Heidelberg}
}

@misc{ruder2017overviewgradientdescentoptimization,
      title={An overview of gradient descent optimization algorithms}, 
      author={Sebastian Ruder},
      year={2017},
      eprint={1609.04747},
      archivePrefix={arXiv},
      primaryClass={cs.LG},
      url={https://arxiv.org/abs/1609.04747}, 
}

@inproceedings{
kostrikov2022offline,
title={Offline Reinforcement Learning with Implicit Q-Learning},
author={Ilya Kostrikov and Ashvin Nair and Sergey Levine},
booktitle={International Conference on Learning Representations},
year={2022},
url={https://openreview.net/forum?id=68n2s9ZJWF8}
}

@misc{geminiroboticsteam2025geminirobotics15pushing,
      title={Gemini Robotics 1.5: Pushing the Frontier of Generalist Robots with Advanced Embodied Reasoning, Thinking, and Motion Transfer}, 
      author={{Gemini Robotics Team}},
      year={2025},
      eprint={2510.03342},
      archivePrefix={arXiv},
      primaryClass={cs.RO},
      url={https://arxiv.org/abs/2510.03342}, 
}

@misc{gpt5,
      title={OpenAI GPT-5 System Card}, 
      author={{OpenAI Team}},
      year={2025},
      eprint={2601.03267},
      archivePrefix={arXiv},
      primaryClass={cs.CL},
      url={https://arxiv.org/abs/2601.03267}, 
}
\bibliographystyle{iclr2026_conference}
\newpage
\appendix
\section{Appendix}
\subsection{Model Architecture and Training Details}
\label{appendix:model}
\begin{figure}[h]
    \centering
    \includegraphics[width=.8\textwidth]{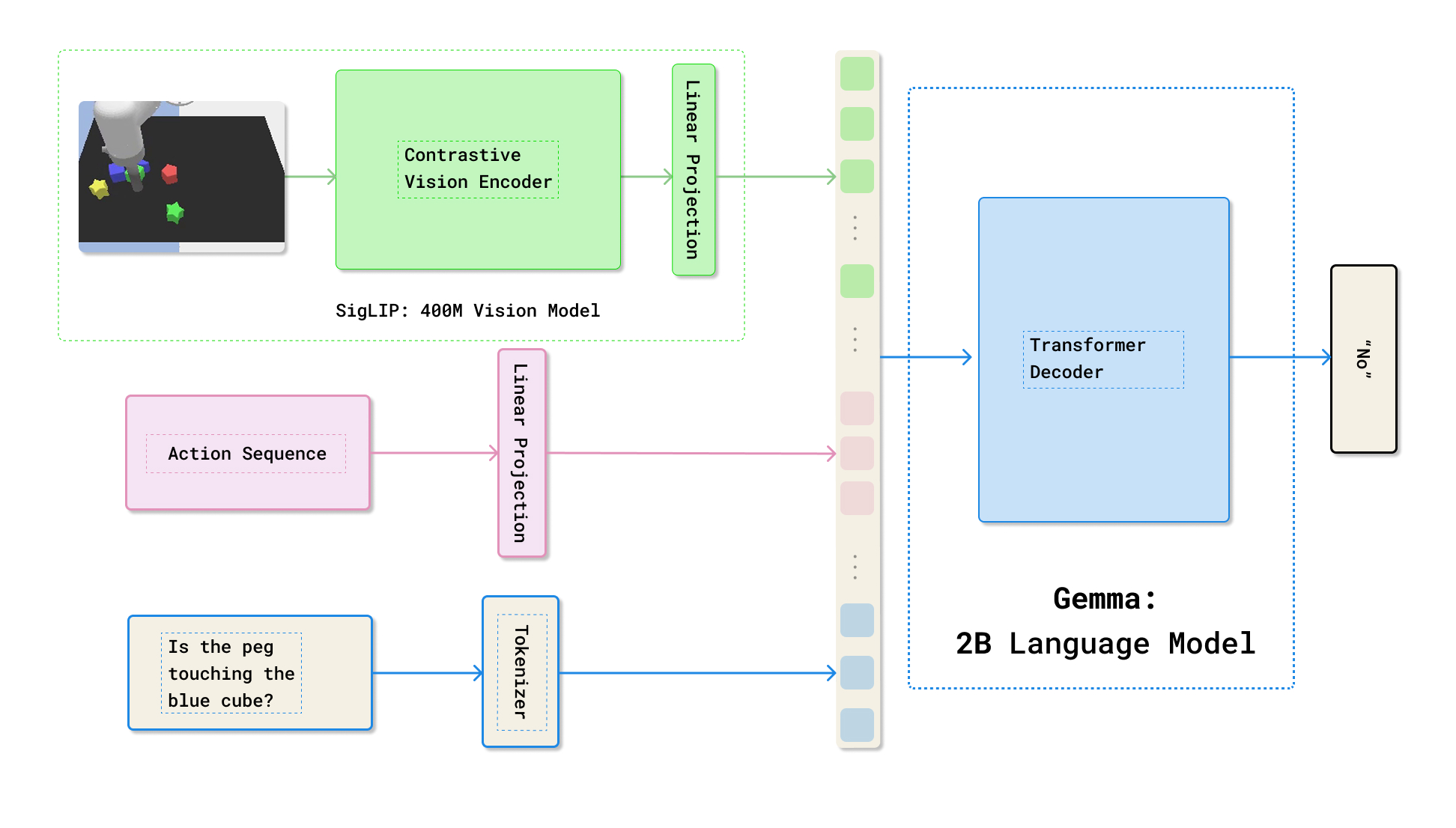}
    \caption{Architecture of \methodfull }
    \label{fig:architecture}
\vspace{-.5em}
\end{figure}
Fig. \ref{fig:architecture} shows the architecture of \algnamefull. We use the Paligemma 3B checkpoint as our base model. The only new component we introduce is a linear projection matrix that is dimension $\texttt{act\_dim} \times \texttt{2048}$ where 2048 is the embed dimension of the Gemma model. We perform full weight fine-tuning on all model parameters using a linear LR decay starting at $1e^{-5}$ for approximately $24,000$ gradient steps on LangTable and $64,000$ gradient steps for OGBench. We use an effective batch size of $96$. Each model is trained on a node comprising 4 AMD Instinct MI250X GPUs (each equipped with 2 MI200 GPU accelerators), resulting in a total training time of approximately 24 hours.

\subsection{Baselines and Hyperparameters}
\label{appendix:baselines}

IDQL \citep{idql} is an offline RL method that applies implicit Q-learning to reweight a behavior diffusion-based policy. We use the base diffusion policy architecture for \method as the policy for IDQL, except with an action horizon of 8 instead of 16.  For the Q and Value functions in IDQL, we only condition on the current observation.

For the AVD baseline, we train a latent action-conditioned transformer video diffusion model, based on the architecture of Unified World Models \citep{zhu2025uwm}, without the action prediction head. Since our semantic world model is trained to answer questions with a variable action horizon, we use SWM to extract semantic information from the predicted future frame by including zero actions.  Due to the computational cost of running the AVD forward and then using the generated frame for VQA, we are unable to run this baseline with a high number of samples. Since the MPPI initial samples were initialized from the base policy, we perform 10 iterations of MPPI with 16 samples to get our final action prediction. Each AVD run takes around 10 hours on a single GPU.

The hyperparameters used for the base diffusion model, the IDQL algorithm, and the AVD model are detailed in Tab. \ref{tab:hyperparamsbl}. The only difference across environments is the size of the input image. All models are trained with the AdamW optimizer \citep{adamw}. 

\subsubsection{Latent World Models}
\label{appendix:dreamer}
We evaluated a latent model-based RL method, Dreamer \citep{Hafner2020Dream}, on our LangTable tasks. Across three offline tasks, we found that it is unable to solve them, achieving an average success rate of $1.33\%$ from the same offline dataset and sparse rewards used for our SWM, IDQL, and AVD methods. We validated our results using the online version of Dreamer with fixed initialization and dense rewards. This variant achieves $100\%$ success, indicating Dreamer's reliance on online exploration data and dense rewards. 

We found similar behavior evaluating the official implementation of TDMPC2 \cite{hansen2024tdmpc}, modified to use the sparse offline dataset. It achieved on average across three language table tasks a $10\%$ success rate and struggled to learn effectively without dense reward signals or active exploration.

\begin{table*}[t]
\centering
\caption{\footnotesize{Hyperparameters for MPPI, IDQL, Diffusion, and AVD Model}}
\label{tab:hyperparamsbl}
\begin{tabular}{ll}
\toprule
\multicolumn{2}{c}{\textbf{Diffusion}} \\
\midrule
Batch size              & 128 \\
Epochs                  & 100 \\
Action horizon          & 16 \\
Observation horizon     & 2 \\
Diffusion iters         & 100 \\
Eval diffusion iters    & 10 \\
Traj end padding (steps)& 12 \\
\midrule
\multicolumn{2}{c}{\textbf{IDQL}} \\
\midrule
Gradient steps          & 250,000 \\
Batch size              & 128 \\
IQL $\tau$              & 0.8 \\
Test time samples       & 1000 \\
Temperature             & 0.5 \\
Discount ($\gamma$)     & 0.99 \\
Critic hidden dim       & 256 \\
Critic learning rate    & 0.0003 \\
Num layers              & 3 \\
\midrule
\multicolumn{2}{c}{\textbf{AVD Model}} \\
\midrule
Embed dim               & 768 \\
Vision backbone         & ViT-B/32 \\
Timestep embed dim      & 512 \\
Latent patch shape      & [2,2,2] \\
Num Transformer Layers  & 12 \\
Num heads               & 12 \\
Train steps             & 1000 \\
Inference steps         & 100    \\
Total steps             & 100,000 \\
Global batch size       & 288 \\
Learning rate           & 1e-4 \\
Weight decay            & 1e-6 \\
\midrule
\multicolumn{2}{c}{\textbf{MPPI}} \\
\midrule
Num Iterations               & 30 \\
Samples         & 192 \\
Temperature      & .2 \\
\bottomrule
\end{tabular}
\end{table*}

\subsection{Environments and Tasks}
\label{appendix:environments}

\subsubsection{Environment Details}
\begin{figure}[t]
    \centering
    \includegraphics[width=.8\textwidth]{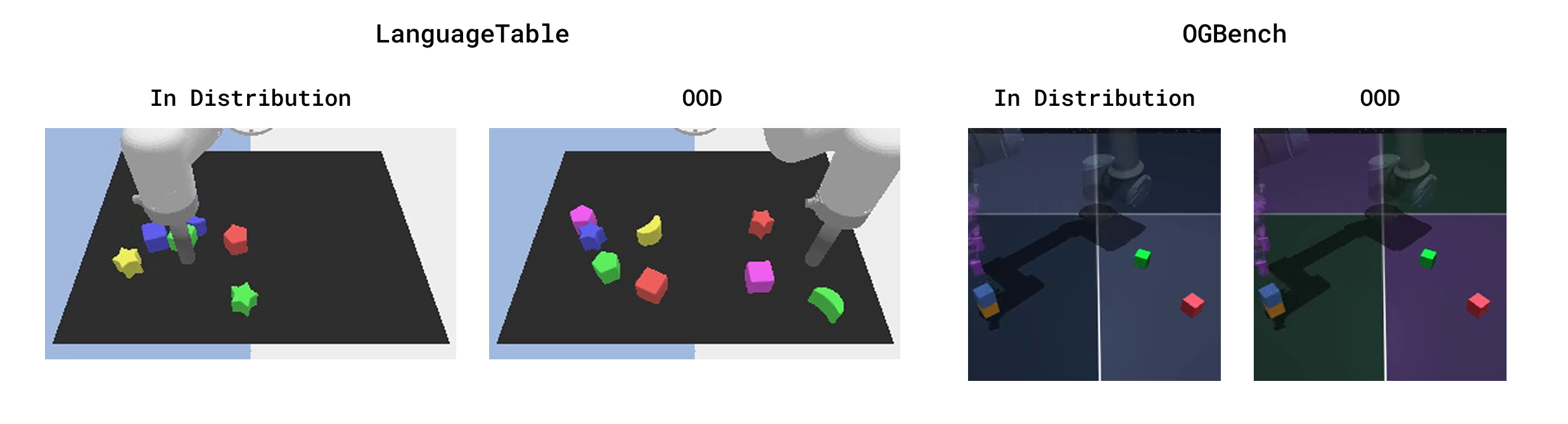}
    \caption{\footnotesize{Out-of-distribution configurations for the evaluation tasks. LangTable is configured to have OOD block/color combinations. OGBench is configured to have a different color background.}}
    \label{fig:ood_figure}
\vspace{-.5em}
\end{figure}
Fig. \ref{fig:envfigure} shows an example of each type of task we used to evaluate \algname. In Fig. \ref{fig:ood_figure}, we provide examples of out-of-distribution configurations used to evaluate the generalization capabilities of \algname{}. More details about each environment and task are discussed below.

\textbf{LangTable} The LangTable environment has a control frequency of 10 Hz. For each task, we terminate each episode after 120 environment steps. Our observation space is a single \texttt{180$\times$320} RGB image of the table.  The action space is xy delta poses, ranging from -.03 to .03.  Our reach block task is marked as a success if the peg made contact with the target block. The separate block task is marked as a success if the L2 distance between the target block and the blocks to separate it from is over $.1$ M. For pushing blocks together, the episode is marked as a success if the L2 distance between the two target blocks is less than $.075$. The expert and noisy demonstrations used for our offline dataset and expert diffusion dataset are collected on environment seeds 0-300, and we evaluate on seeds 6000-6050.  For the \method improvement, we use an action chunk of 8, a gradient learning rate of 0.02, 10 planning iterations, and execute 4 out of the 16 predicted actions before replanning.  We use a gradient clipping of 1 before updating each action during planning.

\textbf{OGBench} We use the cube environment as the basis for our tasks. This environment has a control frequency of 10Hz, and we terminate each episode after 200 steps. Our observation space is a single \texttt{224$\times$224} RGB image.  The action space is 5-dimensional, comprising of delta xyz and orientation, and a gripper action. For the ReachCube task, we measure success as the gripper pads touching the cube. For our cube stacking task, we initialize all block poses randomly and then define success as the first cube being stacked on top of the second cube, with a gap between the top cube and the robotic gripper.  The expert and noisy demonstrations used for our offline dataset and expert diffusion dataset are collected on environment seeds 0-300, and we evaluate on seeds 6000-6050.   For the \method improvement, we use an action chunk of 8, a gradient learning rate of 0.2, 20 planning iterations, and execute 4 out of the 16 predicted actions before replanning. We use gradient clipping of 10 before updating each action during planning.

\subsubsection{Question-Answer Dataset Curation}
\label{appendix:qaprompts}
We precompute the future QA pairs for our offline dataset. For each state, we sample four different action horizon lengths between $0$ and $20$, and generate a set of questions for each sampled horizon. Tab. \ref{tab:qaquestions} shows the question types and an example of each question type on both the LangTable and OGBench environments.

\begin{longtable}{@{}l p{0.65\linewidth}@{}}
\caption{Question types and examples for LangTable and OGBench} \label{tab:qaquestions}\\

\toprule
\textbf{Type} & \textbf{Example} \\
\midrule
\endfirsthead

\multicolumn{2}{c}%
{{\tablename\ \thetable{} -- continued from previous page}} \\
\toprule
\textbf{Type} & \textbf{Example} \\
\midrule
\endhead

\midrule
\multicolumn{2}{r}{{Continued on next page}} \\
\endfoot

\bottomrule
\endlastfoot

\multicolumn{2}{@{}l}{\textbf{LangTable}} \\
Block touching & Is the red star touching the blue cube? \\
Peg to block & Is the green cube next to the peg? \\
Block board position & Is the red star in the center of the board? \\
Peg block relative direction & Is the peg above the red cube block? \\
Block to block relative direction & Is the red star to the right of the blue cube? \\
Block move direction & Did the red cube move left? \\
Block move & Did the red star block move? \\
Peg move direction & Did the robotic peg move downward? \\
Block to block closer & Are the red star and blue cube closer together? \\
Peg to block closer & Is the robotic peg closer to the red cube? \\[0.75em]

\multicolumn{2}{@{}l}{\textbf{OGBench}} \\
Cube grasped & Is the red cube grasped by the robot? \\
Gripper touching block & Is the blue cube touching the robot gripper? \\
Block touching block & Is the green cube touching the yellow cube? \\
Block on top of block & Is the red cube on top of the blue cube? \\
Gripper closer to block & Is the gripper closer to the green cube? \\
Block closer to block & Is the red cube closer to the blue cube? \\
\end{longtable}

For each question type, we also use multiple variations in wording.  
For example, for \emph{block touching} questions, given two blocks \{block1\} and \{block2\}, we use:

\begin{itemize}[leftmargin=2em, itemsep=0.25em, parsep=0em]
    \item Is the \{block1\} touching the \{block2\}?
    \item Are the \{block1\} and \{block2\} blocks in contact with each other?
    \item Is there contact between the \{block1\} block and the \{block2\} block?
    \item Does the \{block1\} touch the \{block2\}?
    \item Is the \{block1\} block in physical contact with the \{block2\} block?
    \item Are the \{block1\} and \{block2\} blocks touching each other?
    \item Is the \{block1\} and \{block2\} directly touching?
    \item Do the \{block1\} and \{block2\} blocks meet?
\end{itemize}

\subsubsection{Task Specification}
For each task, we use a fixed set of questions and answers to specify the goals. All of our tasks are single-subgoal tasks except the stack cube task, which has two goals. In order to create a multi-step task for LangTable, we use two subgoals of independent Block to Block tasks, and use the \method to pick the behavior policy and the subgoal to use.  The questions, answers, and weights for all tasks are in shown Table  \ref{tab:qa_all}. 
\begin{table}[h]
\centering
\caption{QA pairs used for task rewards}
\label{tab:qa_all}
\begin{tabular}{l p{0.5\linewidth} c c}
\toprule
\textbf{Task} & \textbf{Question} & \textbf{Weight} & \textbf{Desired Answer} \\
\midrule

\multirow{2}{*}{Reaching LT} 
 & Is the robotic peg touching the \{target\_block\}? & 0.8 & Yes \\
 & Is the robotic peg closer to the \{target\_block\}? & 0.2 & Yes \\
\midrule

\multirow{2}{*}{Reaching OG} 
 & Is the robotic gripper touching the \{target\_block\}? & 0.8 & Yes \\
 & Is the robotic gripper closer to the \{target\_block\}? & 0.2 & Yes \\
\midrule

\multirow{2}{*}{Separate Blocks} 
 & Is the robotic peg touching the \{center\_block\}? & 0.6 & Yes \\
 & Is the \{avoid block\} touching the \{center block\}? & 0.4 & No \\
\midrule

\multirow{2}{*}{Block to Block} 
 & Is the \{first\_block\} touching the \{second\_block\}? & 0.8 & Yes \\
 & Are the \{first\_block\} and the \{second\_block\} closer together? & 0.2 & Yes \\
\midrule

\multirow{3}{*}{Cube Stacking} 
 & \textbf{Subgoal 1: Pick up the first cube} & & \\
 & Is the robot grasping the \{first\_block\}? & 1.0 & Yes \\
\cmidrule{2-4}
 & \textbf{Subgoal 2: Stack the blocks} & & \\
 & Is the \{first\_block\} on top of the \{second\_block\}? & 0.6 & Yes \\
 & Is the robot grasping the \{first\_block\}? & 0.4 & Yes \\
\bottomrule
\end{tabular}
\end{table}

\subsection{Additional Experiments}

\subsection{Full Improvement Results}
We provide the full improvement results corresponding to Fig. \ref{fig:improvement} in the experiments section.
\begin{table}[t]
    \caption{\footnotesize{\textbf{Improvement Results.} SWM model improvement results on planning tasks in LangTable and OGBench on in-distribution scenes. Reported success rates over $n=50$ seeds with 95\% confidence intervals (normal approximation). The top tasks are LangTable and the bottom tasks are OGBench.}}
    \centering
    \begin{tabular}{lcccc}
    \toprule
    Task & Base Policy & IDQL & AVD & SWM \\
    \midrule
    Push Green Cube to Blue Moon & 6\% $\pm$ 6.6 & 8\% $\pm$ 7.5 & 48\% $\pm$ 13.8 & \textbf{78\%} $\pm$ 11.5\\
    Push Red Moon to Green Star & 18\% $\pm$ 10.6 & 8\% $\pm$ 7.5 & 44\% $\pm$ 13.8 & \textbf{80\%} $\pm$ 11.1\\
    Push Red Pentagon to Blue Moon & 14\% $\pm$ 9.6 & 12\% $\pm$ 9.0 & 38\% $\pm$ 13.5 & \textbf{80\%} $\pm$ 11.1\\
    Push Yellow Pentagon to Red Moon & 18\% $\pm$ 10.6 & 8\% $\pm$ 7.5 & 34\% $\pm$ 13.1 & \textbf{86\%} $\pm$ 9.6\\
    Push Yellow Star to Blue Cube & 16\% $\pm$ 10.2 & 10\% $\pm$ 8.3 & 62\% $\pm$ 13.5 & \textbf{84\%} $\pm$ 10.2\\
    \midrule
    Stack Blue Cube on Yellow Cube & 52\% $\pm$ 13.8 & 8\% $\pm$ 7.5 & 50\% $\pm$ 13.9 & \textbf{82\%} $\pm$ 10.6\\
    Stack Blue Cube on Green Cube & 44\% $\pm$ 13.8 & 16\% $\pm$ 10.2 & 46\% $\pm$ 13.8 & \textbf{84\%} $\pm$ 10.2\\
    Stack Yellow Cube on Red Cube & 40\% $\pm$ 13.6 & 24\% $\pm$ 11.8 & 44\% $\pm$ 13.8 & \textbf{62\%} $\pm$ 13.5\\
    \bottomrule
    \end{tabular}
    \label{tab:improvement_ci}
\end{table}

\subsection{Ablations}
\label{appendix:ablations}
\subsubsection{Ablation on question weights}
We ablate the inclusion of question weights across all of the in-distribution LangTable and OGBench tasks. We found that removing weights decreased performance by an average of $2.4\%$ in LangTable and increased performance by $3.3\%$ in ogbench. Full results are in Tab. \ref{tab:weight_ablation}.

\subsubsection{Ablation on planning without base policy}
To evaluate whether SWM can plan effectively without relying on a base policy, we conducted an ablation study on three LangTable pushing tasks. In this setting, the initial action sequence was sampled uniformly from the environment’s action space, removing any prior structure provided by a base policy. Despite the lack of a warm start, SWM was able to successfully plan under these randomly initialized trajectories, achieving success rates of $46\%$, $50\%$, and $58\%$ on the tasks of pushing the yellow pentagon to the red moon, pushing the red moon to the green star, and pushing the yellow star to the blue cube. These results demonstrate that the gradients are reasonable even from random initializations.

\subsubsection{Ablation on question phrasings}
To evaluate the robustness of SWM to different question phrasings, we conducted an ablation measuring performance under both in-distribution and out-of-distribution question phrasings. For each task, we evaluated SWM using two different phrasings seen during training and two novel OOD phrasings not present in the SAQA dataset. As shown in Table~\ref{tab:phrasing_ablation}, SWM maintains strong performance across all phrasing variants, with only minor drops under OOD formulations. These results demonstrate SWMs robustness to new question phrasings.

\begin{table}[ht]
    \caption{\textbf{Ablation on question weights.} Success rates for SWM with weights vs. SWM without weights on LangTable and OGBench tasks. Reported over $n=50$ seeds with 95\% confidence intervals.}
    \label{tab:weight_ablation}
    \centering
    \begin{tabular}{lcc}
    \toprule
    Task & SWM (with weights) & SWM (no weights) \\
    \midrule
    Push Green Cube to Blue Moon & 78\% $\pm$ 11.5 & 72\% $\pm$ 12.4\\
    Push Red Moon to Green Star & 80\% $\pm$ 11.1 & 78\% $\pm$ 11.4\\
    Push Red Pentagon to Blue Moon & 80\% $\pm$ 11.1 & 82\% $\pm$ 10.6\\
    Push Yellow Pentagon to Red Moon & 86\% $\pm$ 9.6 & 88\% $\pm$ 9.0\\
    Push Yellow Star to Blue Cube & 84\% $\pm$ 10.2 & 76\% $\pm$ 11.8\\
    \midrule
    Stack Blue Cube on Yellow Cube & 82\% $\pm$ 10.6 & 82\% $\pm$ 10.6\\
    Stack Blue Cube on Green Cube & 84\% $\pm$ 10.2 & 78\% $\pm$ 11.4\\
    Stack Yellow Cube on Red Cube & 62\% $\pm$ 13.5 & 78\% $\pm$ 11.4\\
    \bottomrule
    \end{tabular}
\end{table}

\begin{table}[h!]
    \centering
    \caption{\textbf{Ablation on Question Phrasing.} Success rates of SWM under in-distribution (ID) and out-of-distribution (OOD) task phrasings. Reported with 95\% confidence intervals.}
    \begin{tabular}{lccccc}
        \toprule
        Task & Base Policy & ID 1 & ID 2 & OOD 1 & OOD 2 \\
        \midrule
        \small Push Red Moon to Green Star & 18\% $\pm$ 10.6 & 72\% $\pm$ 12.4 & 88\% $\pm$ 9.0 & 84\% $\pm$ 10.2 & 78\% $\pm$ 11.5 \\
        \small Push Yellow Star to Blue Cube & 16\% $\pm$ 10.2 & 78 \% $\pm$ 11.5 & 86\% $\pm$ 9.6 & 84\% $\pm$ 10.2 & 72\% $\pm$ 12.4 \\
        \small Push Yellow Pentagon to Red Moon & 18\% $\pm$ 10.6 & 88\% $\pm$ 9.0& 88\% $\pm$ 9.0 & 86\% $\pm$ 9.6 & 76\% $\pm$ 11.8 \\
        \bottomrule
    \end{tabular}
    \label{tab:phrasing_ablation}
\end{table}

\subsubsection{Automatic question generation}
\label{appendix:autoqa}
To validate the broader applicability of SWM beyond simulators with privileged state information, we evaluate an automatic dataset generation pipeline using a VLM to provide supervision for question answers. We evaluate the accuracy of Gemini Embodied Reasoning \citep{geminiroboticsteam2025geminirobotics15pushing} in providing question-answer supervision on both LangTable and OGBench. In addition, we have measured the accuracy on a small set of manually annotated question-answer pairs from the Droid datasets. Across both settings, Gemini achieved strong accuracy compared to ground-truth answers, indicating that frontier VLMs show a path towards generating the SAQA dataset without relying on oracle information in both sim and real. Detailed results are in Tab. \ref{tab:langtable_saqa}, \ref{tab:ogbench_saqa}, and \ref{tab:droid_saqa}. For each question, we issued multiple queries to the model, counted each response as a vote, and selected the answer with the highest vote count.

\begin{table}[t]
\centering

\begin{minipage}[t]{0.47\linewidth}
\centering
\caption{Gemini-ER Accuracy on LangTable}
\label{tab:langtable_saqa}
\small
\begin{tabular}{lc}
\toprule
\textbf{Question Type} & \textbf{Accuracy} \\
\midrule
block\_touching & 0.93 \\
peg\_to\_block   & 0.89 \\
block\_closer   & 0.88 \\
\midrule
\textbf{Average} & \textbf{0.90} \\
\bottomrule
\end{tabular}
\end{minipage}
\hfill
\begin{minipage}[t]{0.47\linewidth}
\centering
\caption{Gemini-ER Accuracy on OGBench}
\label{tab:ogbench_saqa}
\small
\begin{tabular}{lc}
\toprule
\textbf{Question Type} & \textbf{Accuracy} \\
\midrule
cube\_grasped          & 1.00 \\
block\_ontop\_block    & 0.95 \\
block\_touching\_block & 0.97 \\
block\_block\_closer   & 0.75 \\
\midrule
\textbf{Average} & \textbf{0.92} \\
\bottomrule
\end{tabular}
\end{minipage}

\vspace{1em}

\begin{minipage}[t]{0.52\linewidth}
\centering
\caption{Gemini-ER Accuracy on DROID}
\label{tab:droid_saqa}
\small
\begin{tabular}{lc}
\toprule
\textbf{Question Type} & \textbf{Accuracy} \\
\midrule
claw\_hold     & 0.89 \\
obj            & 0.91 \\
obj\_relative  & 0.82 \\
\midrule
\textbf{Average} & \textbf{0.87} \\
\bottomrule
\end{tabular}
\end{minipage}

\end{table}

\subsubsection{Automatic Task Decomposition}
We evaluate the feasibility of using a VLM for automatic high-level decomposition. We find that for long-horizon tasks, GPT 5.1 \citep{gpt5} is able to break the task into subtasks and create a set of questions and desired answers to plan with. We observe that the GPT-generated questions were also more varied in phrasing. When paired with SWM planning, these planning parameters perform comparably to our oracle question-answer set planning results. Results are shown in Tab. \ref{tab:llm_plan}

We used structured JSON output with a full prompt text of  ``You are a robotic agent planning to push blocks around on a table. Break your task down into key information and brief, absolute and relative questions. Question examples we've trained on: Is the red star touching the blue cube? Is the green cube next to the peg? Is the red star in the center of the board? Is the peg above the red cube block? Is the red star to the right of the blue cube? Did the red cube move left? Did the red star block move? Did the robotic peg move downward? Are the red star and blue cube closer together? Is the robotic peg closer to the red cube?"

\begin{table}[t]
\centering
\caption{Success rate (\%) on long-horizon LangTable tasks with and without LLM-derived task decompositions over 50 seeds. MS1 - red pentagon to blue moon, yellow pentagon to red moon. MS2 - yellow star to blue cube, yellow pentagon to red moon. MS3 - yellow star to blue cube, red pentagon to blue moon. MS4 - green cube to blue moon, yellow pentagon to red moon.}
\label{tab:llm_plan}
\small
\begin{tabular}{lccc}
\toprule
\textbf{Task} & \textbf{Base Policy} & \textbf{SWM+grad} & \textbf{SWM+grad + LLM planning} \\
\midrule
MS1    & 6  & 50 & 62 \\
MS2     & 4  & 66 & 42 \\
MS3       & 4  & 54 & 52 \\
MS4      & 2  & 54 & 44 \\

\midrule
\textbf{Average} & \textbf{4} & \textbf{56} & \textbf{50} \\
\bottomrule
\end{tabular}
\end{table}

\subsubsection{Visualization of Attention Maps}
\label{appendix:attn}
\begin{figure}
    \centering
    \includegraphics[width=\linewidth]{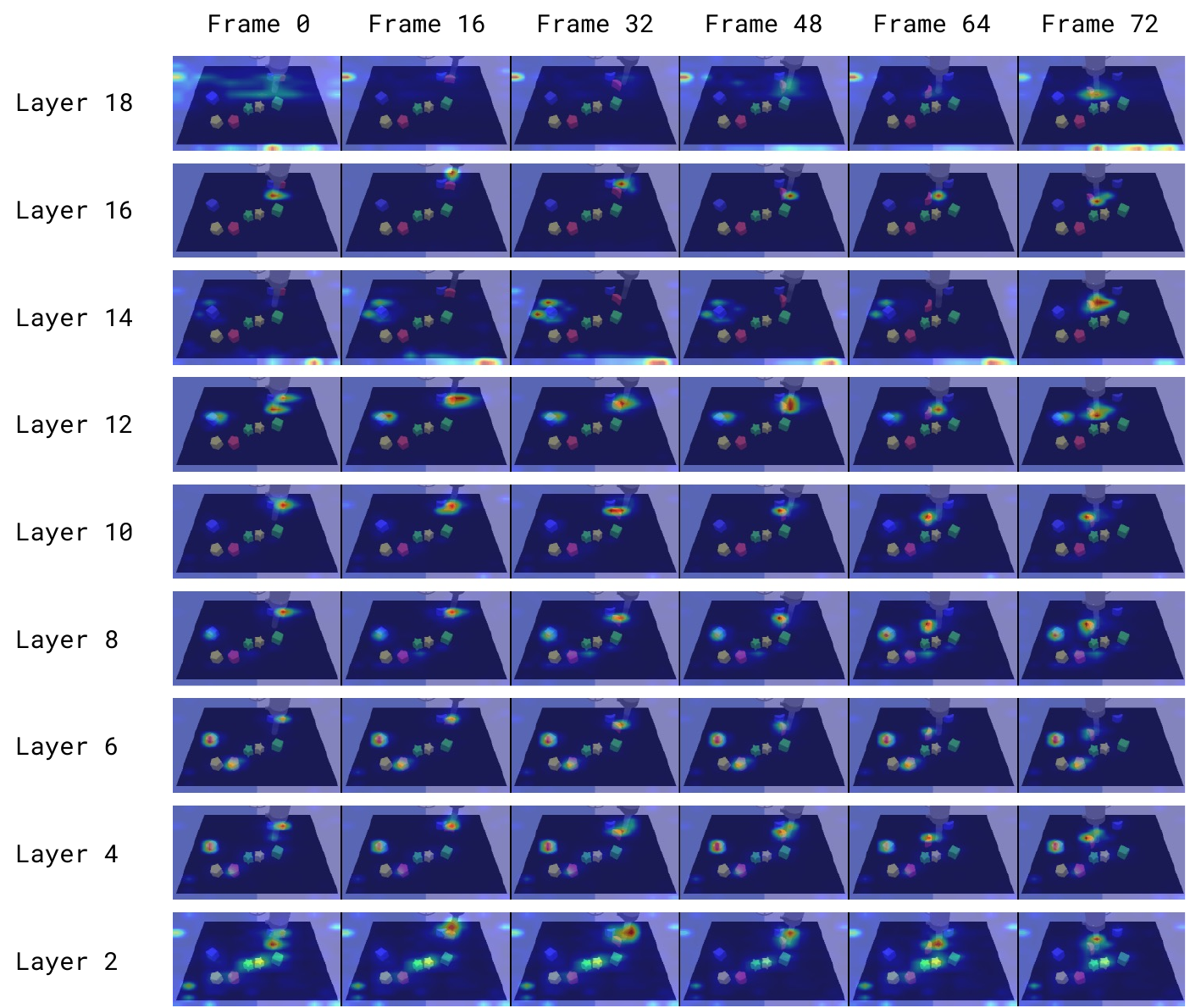}
    \caption{Attention maps in different layers of SWM. Question: ``Is the red moon touching the blue block?''}
    \label{fig:attn-layer}
\end{figure}

\begin{figure}
    \centering
    \includegraphics[width=\linewidth]{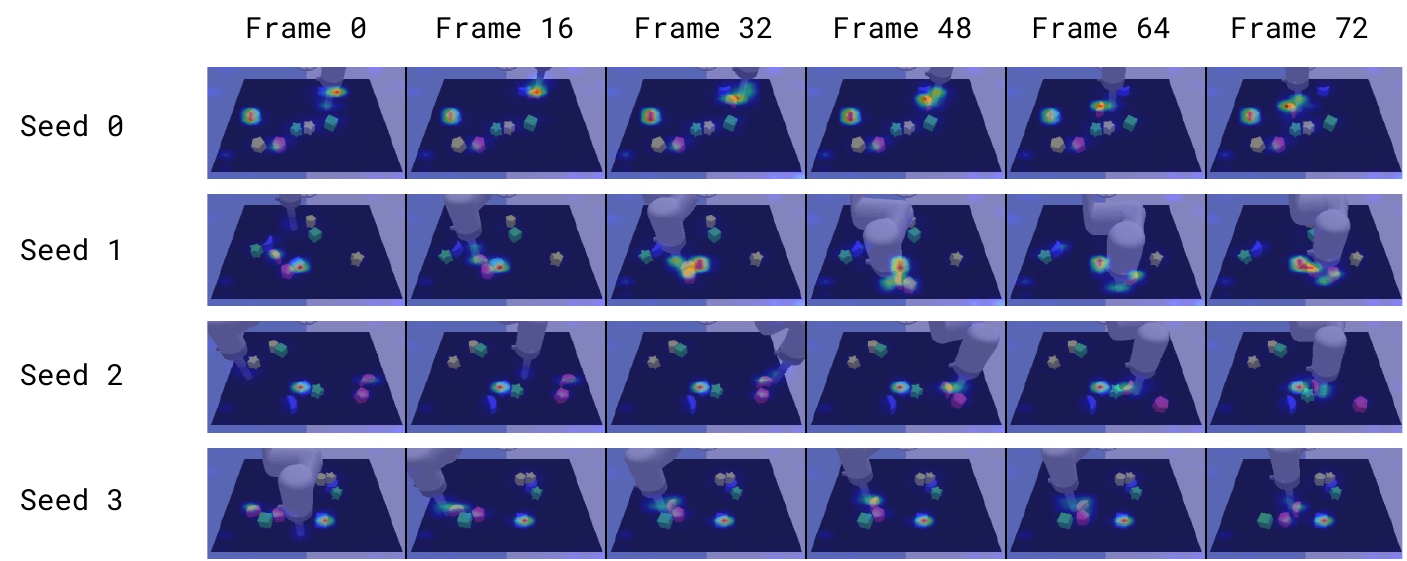}
    \caption{Attention maps for different trajectories. Question: ``Is the red moon touching the blue block?''}
    \label{fig:attn-seed}
\end{figure}

We provide additional visualizations of the attention map. In Fig. \ref{fig:attn-layer}, we visualize the average attention scores from language tokens to image tokens on a consecutive trajectory. We find that different layers capture different semantic information. For example, layers 4 and 6 attend to the red moon and the blue block, whereas later layers also attend to the peg, likely because of the need to reason about the result of actions. In Fig. \ref{fig:attn-seed} we visualize the attention map in layer 4 on different trajectories, showing that the layer consistently attends to the correct objects. 
\subsubsection{Visualization of Gradient-Based Planning}
\label{appendix:visgradplan}
\begin{figure}[b]
    \centering
    \begin{minipage}{.8\textwidth}
        \centering
        \includegraphics[width=\linewidth]{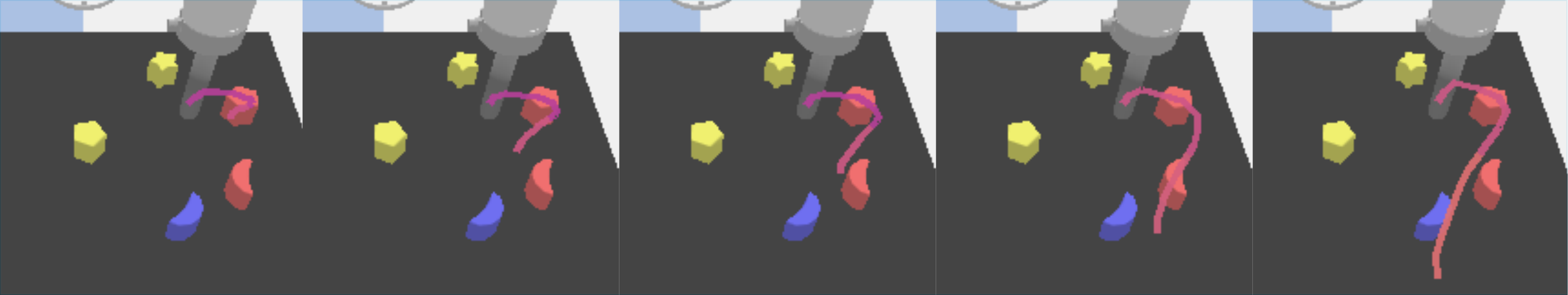}
    \end{minipage}\hfill
    \caption{Visualization of gradient-based planning on the LangTable - Red Pentagon to Blue Moon task. The initially proposed action sequence is on the left, and updates to this action sequence go progressively to the right, approaching the optimal trajectory over successive gradient steps.}
    \label{fig:gradplan}
    \vspace{-1em}
\end{figure}

We visualize the gradient-based planning procedure in Fig. \ref{fig:gradplan}. As planning iteration progresses, the candidate action sequence gradually extends to pushing the red pentagon to the blue moon, approaching the optimal trajectory over successive gradient steps.

\subsubsection{Planning Efficiency}
\label{appendix:planningspeed}
We measure the per-action-chunk planning time of MPPI and our gradient-based method in LangTable on a single GPU shown in Tab.~\ref{tab:planning_speed}. MPPI is configured with eight iterations of 16 samples, and gradient-based planning uses 10 iterations on a single candidate trajectory; both methods use a reward sub-chunk size of 8 and a horizon of 16. MPPI's success rate scales with the number of samples and optimization iterations; for our main planning results in Tab.~\ref{tab:planning_results} we use 30 iterations of 192 samples (Tab.~\ref{tab:hyperparamsbl}), which increases the per-action-chunk cost roughly proportionally.

\begin{table}[h]
    \centering
    \caption{Planning speed comparison between MPPI and gradient-based planning, measured on a single GPU in the LangTable environment.}
    \label{tab:planning_speed}
    \begin{tabular}{l c}
        \toprule
        \textbf{Method} & \textbf{Time per action chunk (s)} \\
        \midrule
        MPPI               & 4.48 \\
        Gradient-based     & 1.56 \\
        \bottomrule
    \end{tabular}
\end{table}

\end{document}